\documentclass{article} 
\usepackage{iclr2027_conference,times}

\usepackage{amsmath,amsfonts,bm}

\def\eqref#1{equation~\ref{#1}}

\def\1{\bm{1}}

\DeclareMathAlphabet{\mathsfit}{\encodingdefault}{\sfdefault}{m}{sl}
\SetMathAlphabet{\mathsfit}{bold}{\encodingdefault}{\sfdefault}{bx}{n}

\usepackage{hyperref}
\usepackage{url}
\usepackage{graphicx}
\usepackage{amsthm}
\usepackage{booktabs}
\usepackage{graphicx}
\usepackage{wrapfig}
\usepackage{multirow}
\usepackage{amssymb}
\newtheorem{theorem}{Theorem}

\newtheorem{proposition}[theorem]{Proposition}

\title{Wave Physics as Representation: Tensor-Structured Neural Reconstruction for Ocean Acoustic Fields\par\vspace{0.7em}}

\author{
	\normalfont
	\makebox[\dimexpr\textwidth-2\tabcolsep\relax][c]{%
		\begin{tabular}{c}
			\textbf{Yifan Sun}$^{1}$ \quad
			\textbf{Panqi Chen}$^{1}$ \quad
			\textbf{Weifeng Shi}$^{1}$ \quad
			\textbf{Xiao Fu}$^{2}$ \quad
			\textbf{Lei Cheng}$^{1}$ \\[0.6em]
			$^{1}$College of Information Science and Electronic Engineering, Zhejiang University \\
			$^{2}$Department of ECE, University of Minnesota - Twin Cities
		\end{tabular}%
	}
}

\definecolor{orange}{RGB}{255,107,0}
\definecolor{green}{RGB}{10,150,10}

\definecolor{purple}{RGB}{128, 0, 128}

\definecolor{Gray}{gray}{0.9}
\usepackage{color, colortbl}
\iclrfinalcopy 
\begin{document}

\maketitle
\lhead{Preprint}

 \begin{abstract}
 	Reconstructing three-dimensional (3D) complex ocean acoustic fields from sparse measurements is challenging due to highly oscillatory wave structures and restricted sensing patterns/routes.
 	Existing learning-based approaches generally fall into two categories.
 	First, purely data-driven models can learn complex field patterns from data, but do not explicitly exploit the underlying wave physics that can potentially enhance interpretability and reduce training costs. 
 	Second, physics-informed methods impose a specified physical equation as a reconstruction constraint, but usually require the environment-dependent parameters to be known, which is difficult to acquire in practice.
 	As a result, practical 3D acoustic field reconstruction that is both physically grounded and robust to uncertain environmental information remains largely unexplored.
 	This work takes an alternative route: instead of solving a fully specified wave equation, it extracts a structured representation from the mathematical form of wave physics and learns its components from data.
 	Starting from the Helmholtz equation and normal mode propagation, we derive a structured spatial–angular tensor representation, termed the \(S/c\) representation. 
 	Under a low-rank specialization, it takes the form of a rank-\((L,L,1)\) block-term tensor decomposition, which connects structured sampling patterns/routes often used in ocean sensing practice to sufficient conditions for field recovery.
 	Furthermore, building on this tensor representation, we propose TIDAL, a neural reconstructor that uses the \(S/c\) representation as an inductive bias for structured decoding and aligns information aggregation with the sampling geometry. The neural learner removes the need for known model parameters required by conventional physics-based recovery methods. 
 	Experiments across three structured sampling geometries show that TIDAL achieves the best reconstruction performance and fastest inference among baselines, with clear benefits from the physics-derived representation under limited data and model capacity.
 \end{abstract}

\section{Introduction}
\label{sec:introduction}
\enlargethispage{\baselineskip}

Accurate reconstruction of three-dimensional (3D) complex ocean acoustic fields from sparse measurements is important for underwater sensing and environmental characterization \citep{zhu2024strategic}.
The reconstruction problem is particularly difficult in realistic ocean environments for two reasons.
First, modal propagation and multipath effects produce highly oscillatory interference patterns, making the field fundamentally different from smooth signals \citep{jensen2011computational}.
Second, ocean observations cannot usually be placed arbitrarily, but are constrained by platform motion and array configurations. This leads to structured sampling patterns such as horizontal and vertical (H/V) transects \citep{wan2009spatial,tang2007shallow}, while many established inverse problem recoverability supports are based on random sampling patterns; see, e.g., \cite{candes2010matrix}.
Recovering the full 3D field therefore requires an effective representation of oscillatory wave structure that is compatible with the geometry of sparse observations.

Existing learning-based approaches mainly follow two directions.
Purely data-driven approaches can capture rich field patterns directly from data~\citep{zhao2024recfno,santos2023development,huang2024diffusionpde}. 
However, their representations are typically generic and do not explicitly exploit the governing wave physics. 
Physics-informed methods, such as physics-informed neural networks (PINNs)~\citep{raissi2019physics}, instead incorporate governing physical equations as constraints. This requires scene-dependent physical parameters to be known, which is often difficult in complex ocean environments. Consequently, practical reconstruction methods that can retain physical structure without relying on accurately specified scene-specific parameters remain largely unexplored.

This motivates a different route: using physics at the representation level. 
The central observation is that the mathematical structure of the governing wave equation already reveals how the physical field can be represented. 
Starting from the Helmholtz equation and the normal-mode representation of underwater propagation \citep{jensen2011computational}, we organize the directional modal solutions along the periodic angular coordinate and apply an angular spectral expansion. 
This yields a structured representation of the field in terms of depth–range spatial components \(S\) modulated by angular factors \(c\), termed the \(S/c\) representation.
Under a low-rank specialization, the \(S/c\) representation connects the physics-derived field structure to rank-\((L,L,1)\) block-term tensor decomposition (LL1) recoverability theory, yielding sufficient conditions for unique recovery from structured H/V observations.

\begin{figure*}[t]
	\centering
	\includegraphics[width=0.94\textwidth]{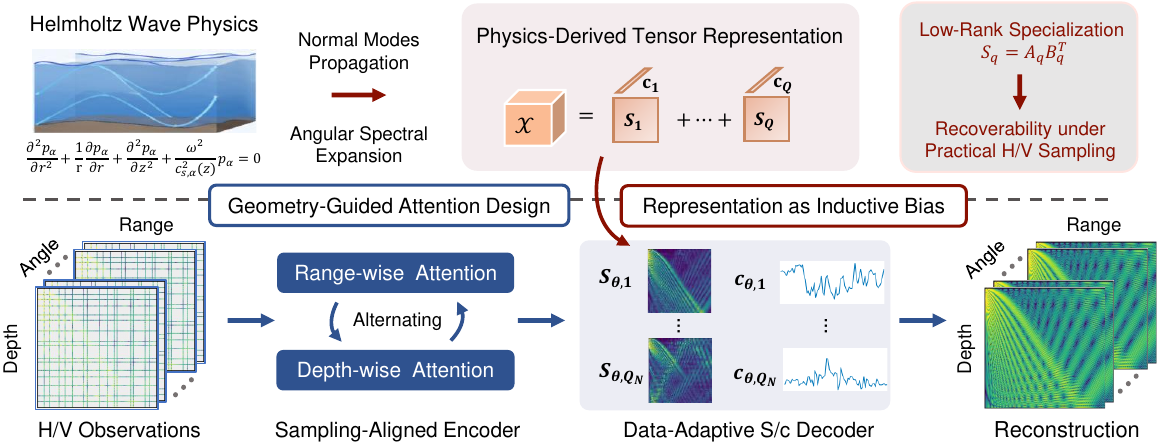}
	\caption{
		Overview of the proposed framework. 
		Starting from the Helmholtz equation, we derive the physics-derived \(S/c\) tensor representation; under a low-rank specialization, it connects H/V observations to LL1 recoverability theory (top). 
		Building on this representation, TIDAL uses the \(S/c\) structure as an inductive bias through data-adaptive decoding, while the H/V observation geometry guides information aggregation through sampling-aligned axial attention (bottom).
	}
	\label{fig:overview}
\end{figure*}

Building on these insights, we propose TIDAL (\underline{T}ensor-\underline{I}nspired \underline{D}ecoding with \underline{A}xial \underline{L}earning), a neural reconstructor guided by both the physics-derived \(S/c\) representation and the structured H/V observation geometry. TIDAL retains the \(S/c\) representation as an inductive bias through structured decoding with data-adaptive components, while sampling-aligned axial attention aggregates information along depth and range following the H/V sensing geometry. In this way, wave physics determines how the field is represented, tensor structure explains when it is recoverable, and attention-based learning reconstructs the missing field from data. Fig.~\ref{fig:overview} summarizes the framework.

Experiments across equispaced, shifted, and jittered H/V sampling geometries show that TIDAL achieves the best reconstruction performance with efficient inference.
Representation analysis shows that the learned spatial components exhibit pronounced singular value decay, while prescribed moderate ranks can limit representation capacity, motivating data-adaptive spatial modeling.
Paired comparisons with direct-output decoding further support the \(S/c\) representation as an effective inductive bias, with particularly clear benefits under limited training data and model capacity.

Our main contributions are summarized as follows:
\begin{itemize}
	
	\item We show that the mathematical form of wave physics directly reveals a structured $S/c$ representation of 3D complex acoustic fields, without requiring exact physical parameters. Under a low-rank specialization, this representation takes the form of LL1, connecting structured H/V observations to classical tensor recoverability theory and yielding sufficient conditions for unique full-field recovery.
	
	\item Building on this representation, we propose TIDAL, which incorporates the physics-derived $S/c$ structure as an inductive bias through structured decoding with data-adaptive components. The H/V observation geometry further guides information aggregation through sampling-aligned axial attention along depth and range.
	
	\item Across three H/V sampling geometries, TIDAL achieves the best reconstruction performance and fastest inference among the baselines, while analysis further highlights the benefits of the physics-derived \(S/c\) representation under limited data and model capacity.
	
\end{itemize}

\section{Related Work}
\label{sec:related_work}
\paragraph{Tensor and low-rank reconstruction.}
Low-rank tensor models, including Canonical polyadic decomposition (CPD) \citep{harshman1970foundations}, Tucker decomposition \citep{tucker1966some}, and block-term decomposition (BTD) \citep{de2008decompositions}, have been widely studied for representing and recovering high-dimensional data from incomplete observations.
Coupled CP and LL1/BTD models have been studied for various applications; see, e.g., recoverability under structured degradation in hyperspectral super-resolution~\citep{sorensen2015coupled,kanatsoulis2018hyperspectral,ding2021hyperspectral}.
Most closely related, real-valued coupled LL1 recovery from structured slab and fiber observations has been studied for spectrum cartography \citep{zhang2020spectrum}. In contrast, our $S/c$ representation is derived from underwater wave physics, and its low-rank specialization connects complex acoustic field reconstruction to existing LL1 recoverability results under structured H/V observations.

\paragraph{Learning-based field reconstruction.}
Purely data-driven approaches have explored neural operators, attention-based models, and generative methods for reconstructing physical fields from incomplete observations~\citep{zhao2024recfno,Tiwari2025latent,santos2023development,huang2024diffusionpde,ma2025physense}. Physics-informed methods instead incorporate governing PDEs into learning through residual-based objectives~\citep{raissi2019physics}, including Helmholtz-equation constraints for acoustic-field reconstruction~\citep{chen2025frequency}. TIDAL differs by incorporating the $S/c$ representation revealed by wave physics through structured decoding and using sampling-aligned attention to match the H/V observation geometry.

\section{Problem Formulation}
\label{sec:problem_formulation}

We consider reconstruction of a 3D complex ocean acoustic field from sparse measurements.
At a fixed frequency, let $p(z,r,\alpha)\in\mathbb{C}$ denote the complex acoustic pressure at depth $z$, range $r$, and angular direction $\alpha$.
Such fields characterize acoustic propagation and interference across space and direction, providing useful information for underwater sensing and environmental characterization \citep{zhu2024strategic}.
In practice, dense spatial sampling is often infeasible because measurements are constrained by platform trajectories and array configurations, naturally leading to structured acquisition patterns such as H/V transects \citep{wan2009spatial,tang2007shallow}.


Classical sparse-field reconstruction typically formulates this task as an underdetermined inverse problem, $\mathbf{y}=\mathbf{A}\mathbf{x}$, where $\mathbf{x}$ is the vectorized field, $\mathbf{y}$ is the measurements, and $\mathbf{A}$ is the sensing matrix.
Recovering $\mathbf{x}$ from $\mathbf{y}$ therefore relies on additional priors, including spatial or sparse representations \citep{sun2017underwater,sun2020kriged}, learned representations \citep{chen2026physics}, and physics-based constraints or dictionaries \citep{chen2025frequency,damiano2025zero}.
However, the inverse formulation alone does not characterize recoverability under the structured H/V sampling geometry considered here.
Such guarantees require the observation geometry to be analyzed jointly with a compatible field representation, as in coupled tensor recovery under systematic slab sampling \citep{zhang2020spectrum}.
We therefore formulate both the field and its structured H/V observations in tensor form.
We discretize the field over $I$ depths $\{z_i\}_{i=1}^{I}$, $J$ ranges $\{r_j\}_{j=1}^{J}$, and $K$ angular directions $\{\alpha_k\}_{k=1}^{K}$, yielding
\begin{equation}
	\mathcal{X}\in\mathbb{C}^{I\times J\times K},
	\qquad
	\mathcal{X}_{ijk}=p(z_i,r_j,\alpha_k).
	\label{eq:field_tensor}
\end{equation}
Let $\mathcal{I}_H\subseteq[I]$ and $\mathcal{J}_V\subseteq[J]$ index the selected horizontal and vertical transects, respectively, where $[I]=\{1,...,I\}$ and similarly for $[J]$ and $[K]$.
The resulting observation set is
\begin{equation}
	\Omega_{\mathrm{HV}}
	=
	\Big[
	(\mathcal{I}_H\times[J])
	\cup
	([I]\times\mathcal{J}_V)
	\Big]
	\times[K].
	\label{eq:hv_sampling}
\end{equation}
The same depth--range sampling pattern is shared across angular directions.
The observed field is therefore
\begin{equation}
	\mathcal{X}_{\mathrm{obs}}
	=
	\mathcal{P}_{\Omega_{\mathrm{HV}}}(\mathcal{X}),
	\label{eq:observation_model}
\end{equation}
where $\mathcal{P}_{\Omega_{\mathrm{HV}}}$ denotes the sampling operator associated with $\Omega_{\mathrm{HV}}$.

Our objective is to recover the full complex tensor $\mathcal{X}$ from $\mathcal{X}_{\mathrm{obs}}$.
We therefore view sparse H/V acoustic-field reconstruction as a structured tensor inverse problem.
In Sec.~\ref{sec:theory}, we derive a structured representation of $\mathcal{X}$ from the underlying wave physics and show how its low-rank specialization connects the H/V observation model to existing LL1 recoverability results.

\section{From Wave Physics to Structured Tensor Representation}
\label{sec:theory}
Following the physics-to-tensor pathway in Fig.~\ref{fig:overview}, we first derive a structured tensor representation of the multi-angle acoustic field from the classical $N\times2$D description of underwater propagation \citep{jensen2011computational}.
We then show that a low-rank specialization takes the form of an LL1 decomposition, which connects the structured H/V observation model to existing recoverability results.

\subsection{From Modal Propagation to the $S/c$ Representation}
\label{sec:physics_btd}

Under the classical $N\times2$D approximation, a 3D acoustic field is represented by a collection of 2D depth–range fields, one for each angular direction $\alpha$, with $p_{\alpha}(z,r)\equiv p(z,r,\alpha)$ \citep{jensen2011computational}.
For time-harmonic propagation in a range-independent waveguide, each directional field satisfies
\begin{equation}
	\frac{\partial^2 p_{\alpha}}{\partial r^2}
	+
	\frac{1}{r}\frac{\partial p_{\alpha}}{\partial r}
	+
	\frac{\partial^2 p_{\alpha}}{\partial z^2}
	+
	\frac{\omega^2}{c_{\mathrm{s},\alpha}^{2}(z)}
	p_{\alpha}
	=0,
	\label{eq:helmholtz}
\end{equation}
where $\omega$ is the angular frequency and $c_{\mathrm{s},\alpha}(z)$ is the sound-speed profile along direction $\alpha$.

To expose the depth–range structure of each directional field, we apply separation of variables, $p_{\alpha}(z,r)=\psi(z)\phi(r)$, reducing \eqref{eq:helmholtz} to the standard depth eigenvalue problem and its associated range equation \citep{jensen2011computational}.
The resulting depth eigenfunctions define the propagation modes. Retaining $L_m$ modes gives
\begin{equation}
	p_{\alpha}(z,r)
	\approx
	\sum_{\ell=1}^{L_m}
	\psi_{\ell,\alpha}(z)\phi_{\ell,\alpha}(r),
	\label{eq:normal_mode}
\end{equation}
where $\psi_{\ell,\alpha}$ is the depth eigenfunction of the $\ell$th mode and $\phi_{\ell,\alpha}$ is its corresponding range-dependent response.
The separated modal equations and boundary conditions are provided in Appendix~\ref{app:angular_approximation}.

We next organize these directional solutions along the angular coordinate.
Since the angular coordinate $\alpha \in [0,2\pi)$ is periodic, the multi-angle field admits the Fourier-series representation \citep{stein2011fourier}
\begin{equation}
	p(z,r,\alpha)
	=
	\sum_{n\in\mathbb{Z}}
	\widehat{p}_{n}(z,r)e^{\mathrm{i}n\alpha}.
	\label{eq:angular_spectral_expansion}
\end{equation}
Here, $n$ denotes the angular Fourier index, and the corresponding depth--range coefficient is
\begin{equation}
	\begin{aligned}
		\widehat{p}_{n}(z,r)
		=
		\frac{1}{2\pi}
		\int_{0}^{2\pi}
		p(z,r,\alpha)e^{-\mathrm{i}n\alpha}\,d\alpha
		\approx
		\frac{1}{2\pi}
		\sum_{\ell=1}^{L_m}
		\int_{0}^{2\pi}
		\psi_{\ell,\alpha}(z)
		\phi_{\ell,\alpha}(r)
		e^{-\mathrm{i}n\alpha}\,d\alpha .
	\end{aligned}
	\label{eq:modal_angular_coefficient}
\end{equation}
Thus, each angular Fourier coefficient aggregates the angle-dependent modal contributions into a depth--range component.
Retaining angular orders $|n|\leq N_{\alpha}$, with $Q=2 N_{\alpha} + 1$, gives the angularly truncated field approximation
\begin{equation}
	p_{N_{\alpha}}(z,r,\alpha)
	=
	\sum_{|n|\leq N_{\alpha}}
	\widehat{p}_{n}(z,r)e^{\mathrm{i}n\alpha}
	=
	\sum_{q=1}^{Q}
	S_q(z,r)c_q(\alpha),
	\label{eq:continuous_sc_representation}
\end{equation}
where \(n_q=-N_\alpha+q-1\), \(c_q(\alpha)\triangleq e^{in_q\alpha}\) is the \(q\)th retained angular Fourier harmonic, and \(S_q(z,r)\triangleq \widehat p_{n_q}(z,r)\) is the corresponding depth–range coefficient field.

\begin{proposition}[Finite-dimensional $S/c$ representation]
	\label{prop:finite_sc_representation}
	Under the $L_m$-mode approximation and angular truncation $|n|\leq N_{\alpha}$, with $Q=2N_\alpha +1$, the depth--range components
	$S_q\in\mathbb{C}^{I\times J}$ and angular factors
	$c_q\in\mathbb{C}^{K}$ defined above yield the sampled field
	\begin{equation}
			\mathcal{X}
			=
			\sum_{q=1}^{Q}
			S_q\circ c_q
			+
			\mathcal{E},
			\label{eq:sc_representation}
		\end{equation}
	where $\circ$ denotes the outer product. The residual $\mathcal{E}$ consists of the discarded modal contribution and the angular Fourier truncation error; if the angular dependence has $s$ square-integrable derivatives, the latter decays as $\mathcal{O}(N_{\alpha}^{-s})$.
\end{proposition}

We refer to this form as the $S/c$ representation. The detailed modal and angular truncation derivations are provided in Appendix~\ref{app:angular_approximation}.

\subsection{Low-Rank Specialization and H/V Recoverability}
\label{sec:hv_recoverability}

For theoretical characterization, we consider a low-rank specialization of the spatial components motivated by the normal-mode structure.
For each sampled direction $\alpha_k$, define $X_k=[p_{\alpha_k}(z_i,r_j)]_{i,j}$, together with the sampled depth-mode and range-response matrices
$\Psi_k=[\psi_{\ell,\alpha_k}(z_i)]_{i,\ell}$,
$\Phi_k=[\phi_{\ell,\alpha_k}(r_j)]_{j,\ell}$.
\eqref{eq:normal_mode} then gives
$X_k\approx\Psi_k\Phi_k^\top$, so each directional slice is generated by at most \(L_m\) retained modal contributions.
Across angle, these modal contributions are not independent: directional variations of the same propagation modes can remain strongly correlated. Consequently, the angular Fourier coefficient fields \(S_q\) in \eqref{eq:sc_representation} can exhibit low effective spatial rank. Motivated by this structure, we consider the specialization
\begin{equation}
	\operatorname{rank}(S_q)\le L,\qquad q=1,\ldots,Q,
\end{equation}
where \(L\) denotes the prescribed spatial rank of each component. 
In the rank-\(L\) case, each spatial component admits a factorization $S_q=A_qB_q^\top$, where $A_q\in\mathbb{C}^{I\times L}$ and $B_q\in\mathbb{C}^{J\times L}$.
For the theoretical analysis, we consider the exact representation case \(\mathcal E=0\). The tensor model becomes
\begin{equation}
	\mathcal{X}
	=
	\sum_{q=1}^{Q}
	(A_qB_q^\top)\circ c_q,
	\label{eq:low_rank_sc_model}
\end{equation}
which takes the form of an LL1 decomposition \citep{de2008decompositions,rontogiannis2021block}.

To relate this LL1 specialization to the H/V observation model, let $P_H$ and $Q_V$ denote the coordinate-selection matrices for the H/V transects defined in Sec.~\ref{sec:problem_formulation}, with
$I_H=|\mathcal{I}_H|$ and $J_V=|\mathcal{J}_V|$.
Applying these coordinate selections preserves the LL1 structure:
\begin{equation}
	\mathcal{Y}^{H}
	=
	\sum_{q=1}^{Q}(P_HS_q)\circ c_q,
	\qquad
	\mathcal{Y}^{V}
	=
	\sum_{q=1}^{Q}(S_qQ_V^\top)\circ c_q,
	\label{eq:hv_ll1_observations}
\end{equation}
and form two coupled LL1 tensors sharing the angular factors $\{c_q\}_{q=1}^{Q}$.
The H/V observation model therefore constitutes a coupled slab-sampling instance of the framework studied by \cite{zhang2020spectrum}, allowing its recoverability result to be specialized to the present setting.
\begin{theorem}
	\label{thm:hv_recoverability}
	\citep{zhang2020spectrum}.
	Assume that the rank-$(L,L,1)$ LL1 factors are generic and that $K\geq Q$.
	If
	\begin{equation}
		I_H\geq 2L,
		\qquad
		J\geq LQ,
		\qquad
		\min\!\left(
		\left\lfloor\frac{I_H}{L}\right\rfloor,Q
		\right)
		+
		\min\!\left(
		\left\lfloor\frac{J}{L}\right\rfloor,Q
		\right)
		\geq Q+2,
		\label{eq:h_recoverability_condition}
	\end{equation}
	then the coupled slab observations identify the block terms \(\{S_q,c_q\}_{q=1}^{Q}\) up to a common permutation and reciprocal nonzero scaling.
	The symmetric condition obtained by interchanging the depth and range modes is given in Appendix~\ref{app:hv_recoverability}. 
\end{theorem}

These ambiguities leave each product \(S_q\circ c_q\) unchanged and therefore do not affect the recomposed tensor. Consequently, the full tensor \(\mathcal X\) is uniquely recoverable from the coupled H/V observations.
Although the result of \citet{zhang2020spectrum} is stated for real-valued LL1 factors, the required LL1 essential-uniqueness property also holds over $\mathbb{C}$ \citep[Theorem~4.4]{de2008decompositions}, so the same coupled slab-sampling argument yields recoverability for the complex-valued acoustic field. Details are given in Appendix~\ref{app:hv_recoverability}.

This result is relevant to ocean sensing because measurements are often constrained to structured transects rather than freely placed spatial samples.
The physics-derived $S/c$ representation therefore converts H/V acquisition into a coupled slab-sampling problem, enabling existing LL1 theory to characterize sufficient conditions for full-field recovery.

\section{TIDAL: Physics-Inspired Neural Reconstruction}
\label{sec:method}

The $S/c$ representation derived in \eqref{eq:sc_representation} provides a physics-grounded structure of the field into depth--range spatial components and angular factors. 
Nonetheless, in addition to the low-rank structure of $S_q$ that mostly models global ``slow variation'' of the field, in realistic propagation settings, acoustic fields can exhibit rich interference patterns and spatial variations.
This motivates more detail-capturing, data-adaptive spatial modeling.
We therefore propose TIDAL, with a decoder that retains the $S/c$ representation while learning data-adaptive spatial and angular components.
To infer these components from sparse H/V observations, TIDAL uses sampling-aligned axial attention to aggregate information along depth and range, following the observation geometry.
Component-wise scale normalization further provides a consistent magnitude convention for the reciprocal scaling ambiguity of the $S/c$ representation.

Given the sparse complex observation
$\mathcal{X}_{\mathrm{obs}}
=
\mathcal{P}_{\Omega_{\mathrm{HV}}}(\mathcal{X})$,
we separate its real and imaginary parts and concatenate them with the spatial observation mask and normalized depth and range coordinates:
\begin{equation}
	Z_{\mathrm{in}}
	=
	\mathrm{Concat}
	\left[
	\Re(\mathcal{X}_{\mathrm{obs}}),
	\Im(\mathcal{X}_{\mathrm{obs}}),
	M,
	Z,
	R
	\right],
	\label{eq:network_input}
\end{equation}
where $M$ denotes the spatial sampling mask, and $Z$ and $R$ are the coordinate maps.
The angular responses are represented along the channel dimension, while the full $I\times J$ depth--range grid is retained throughout the encoder.
A pointwise lifting layer maps the input to an initial latent field
$F^{(0)}\in\mathbb{R}^{I\times J\times D}$, where $D$ denotes the latent feature dimension.

\subsection{Sampling-Aligned Axial Attention}
\label{sec:axial_attention}

The H/V sensing geometry is naturally aligned with the depth and range axes. We therefore factorize spatial attention along these two directions, alternating range- and depth-wise self-attention rather than attending globally over the full spatial grid \citep{ho2019axial}. This geometry-aligned design enables long-range feature interaction while directly matching information propagation to the observation structure.

Given a latent feature field $F^{(\ell)}$, each axial block first applies self-attention independently along the range axis for each depth index,
\begin{equation}
	\widetilde{F}^{(\ell)}
	=
	\mathcal{A}_{r}
	\left(F^{(\ell)}\right),
	\label{eq:range_attention}
\end{equation}
followed by attention along the depth axis for each range index,
\begin{equation}
	F^{(\ell+1)}
	=
	\mathcal{A}_{z}
	\left(\widetilde{F}^{(\ell)}\right).
	\label{eq:depth_attention}
\end{equation}
Here, $\mathcal{A}_{r}$ and $\mathcal{A}_{z}$ denote multi-head self-attention blocks with residual feed-forward updates along the corresponding axes.
Alternating the two operations progressively propagates information from the observed transects across the depth--range field.

\subsection{$S/c$ Decoder}
\label{sec:sc_decoder}

Let
$F\in\mathbb{R}^{I\times J\times D}$
denote the output of the axial encoder, and \(Q_N\) the number of learned $S/c$ components.
The $S/c$ representation in \eqref{eq:sc_representation} organizes the field into spatial components and their corresponding angular factors.
Rather than directly predicting the $K$ complex field slices, the decoder parameterizes the output through separate spatial components and angular factors, which are recombined to reconstruct the full field.

\paragraph{Spatial components.}
The spatial branch directly predicts $Q_N$ dense complex components,
\begin{equation}
	\{S_{\theta,q}\}_{q=1}^{Q_N}
	=
	\mathcal{D}_{S}(F),
	\qquad
	S_{\theta,q}\in\mathbb{C}^{I\times J},
	\label{eq:spatial_head}
\end{equation}
where $\theta$ denotes the learnable model parameters, and $\mathcal D_S$ denotes the spatial decoder heads. The dense parameterization follows the spatial components structure in \eqref{eq:sc_representation} while allowing them to adapt to richer field structure.

\paragraph{Angular factors.}
Following the Fourier parameterization in Sec. \ref{sec:physics_btd}, we encode the angular coordinate with Fourier features and further learn data-adaptive angular factors to capture field-dependent directional structure.
We first aggregate the encoder output by global average pooling (GAP) over the depth--range dimensions,
\begin{equation}
	g
	=
	\operatorname{GAP}(F)\in \mathbb{R}^D,
	\label{eq:global_feature}
\end{equation}
and, for each angle $\alpha_k$, predict
\begin{equation}
	\{c_{\theta,q}(k)\}_{q=1}^{Q_N}
	=
	\mathcal{D}_{c}
	\left(
	g,
	\gamma(\alpha_k)
	\right),
	\label{eq:angular_head}
\end{equation}
where $\mathcal D_c$ denotes the angular decoder heads, and $\gamma(\alpha_k)$ is a Fourier encoding of the angular coordinate \citep{tancik2020fourier}.
The pooled feature provides field-level context, while $\gamma(\alpha_k)$ encodes periodic angular dependence.

\paragraph{Scale normalization.}
The $S/c$ recomposition is unchanged under the reciprocal rescaling
$S_{\theta,q}\!\rightarrow\!\alpha_q S_{\theta,q}$ and
$c_{\theta,q}\!\rightarrow\!\alpha_q^{-1}c_{\theta,q}$~\citep{kolda2009tensor}.
This symmetry allows the same component contribution to be represented with arbitrary paired magnitudes in $S_{\theta,q}$ and $c_{\theta,q}$.
We therefore normalize the spatial magnitude of each component before recomposition.
Specifically, we define
\begin{equation}
	\rho_q
	=
	\left(
	\frac{1}{IJ}
	\sum_{i=1}^{I}
	\sum_{j=1}^{J}
	\left|
	S_{\theta,q}(i,j)
	\right|^2
	\right)^{1/2},
	\label{eq:sc_scale}
\end{equation}
and apply the reciprocal rescaling $\overline{S}_{\theta,q}  = \frac{S_{\theta,q}}{\rho_q},\overline{c}_{\theta,q}=\rho_q c_{\theta,q}.$
The reconstructed field is then
\begin{equation}
	\widehat{\mathcal{X}}_{ijk}
	=
	\sum_{q=1}^{Q_N}
	\overline{S}_{\theta,q}(i,j)
	\overline{c}_{\theta,q}(k).
	\label{eq:sc_reconstruction}
\end{equation}

\subsection{Training with Sampling-Geometry Augmentation}
\label{sec:training}

To reduce sensitivity to a particular transect layout, we resample the observation mask during training.
The numbers of H/V transects are independently selected from a predefined set, and their locations are drawn from equispaced, shifted, and jittered sampling geometries.

The model is trained end-to-end with a complex-valued reconstruction objective combining full-field and region-specific reconstruction errors.
A mild spatial smoothness regularizer is applied to the learned spatial components.
Detailed loss definitions and optimization settings are provided in the Appendix~\ref{app:implementation}.

\section{Experiments}
\label{sec:experiments}

\subsection{Experimental Setup}
\label{sec:experimental_setup}


\paragraph{Dataset.}
The 3D complex acoustic fields were obtained using KRAKEN \citep{porter1992kraken}, a physics-based propagation solver, under realistic ocean environments. The environments include temperature and salinity fields from finite volume coastal ocean model (FVCOM) reanalysis data \citep{chen2006finite}, and bathymetry from the ETOPO1 Global Relief Model \citep{amante2009etopo1}, covering a region spanning approximately \(17.2^\circ\!-\!17.6^\circ\mathrm{N}\) and \(110.1^\circ\!-\!110.5^\circ\mathrm{E}\). We select three source locations and sample 72 azimuthal directions at each location, yielding direction-dependent sound speed profiles and water depths. Combined with three acoustic frequencies (100, 150, 200 Hz) and four source depths (30, 50, 80, 120 m), these configurations span diverse source and propagation conditions. The resulting 660 fields are split into 500/80/80 for training, validation, and testing, with each sample represented on a \(200\times200\times72\) depth–range–azimuth grid.

\paragraph{Baselines.}
We compare against representative learning-based reconstruction methods, including MxT~\citep{chen2024mamba}, LaMO~\citep{Tiwari2025latent}, DiffusionPDE~\citep{huang2024diffusionpde}, and SCPNet~\citep{xia2023scpnet}, using identical acoustic field splits and H/V observation protocols.
Architecture-specific adaptations are detailed in Appendix \ref{app:baseline_implementation}.

\paragraph{Evaluation metrics.}
We report relative root-mean-square error (RMSE) and modal assurance criterion (MAC) \citep{schmid2021spatial} over the full tensor and over unobserved spatial locations. MAC provides a structure-aware measure of normalized similarity between complex acoustic fields, complementing the pointwise reconstruction error measured by RMSE. Unless otherwise specified, evaluation uses H15/V15 (15 horizontal and 15 vertical transects) observations under equispaced, shifted, and jittered geometries.

\begin{figure*}[t]
	\centering
	\includegraphics[width=0.9\textwidth]
	{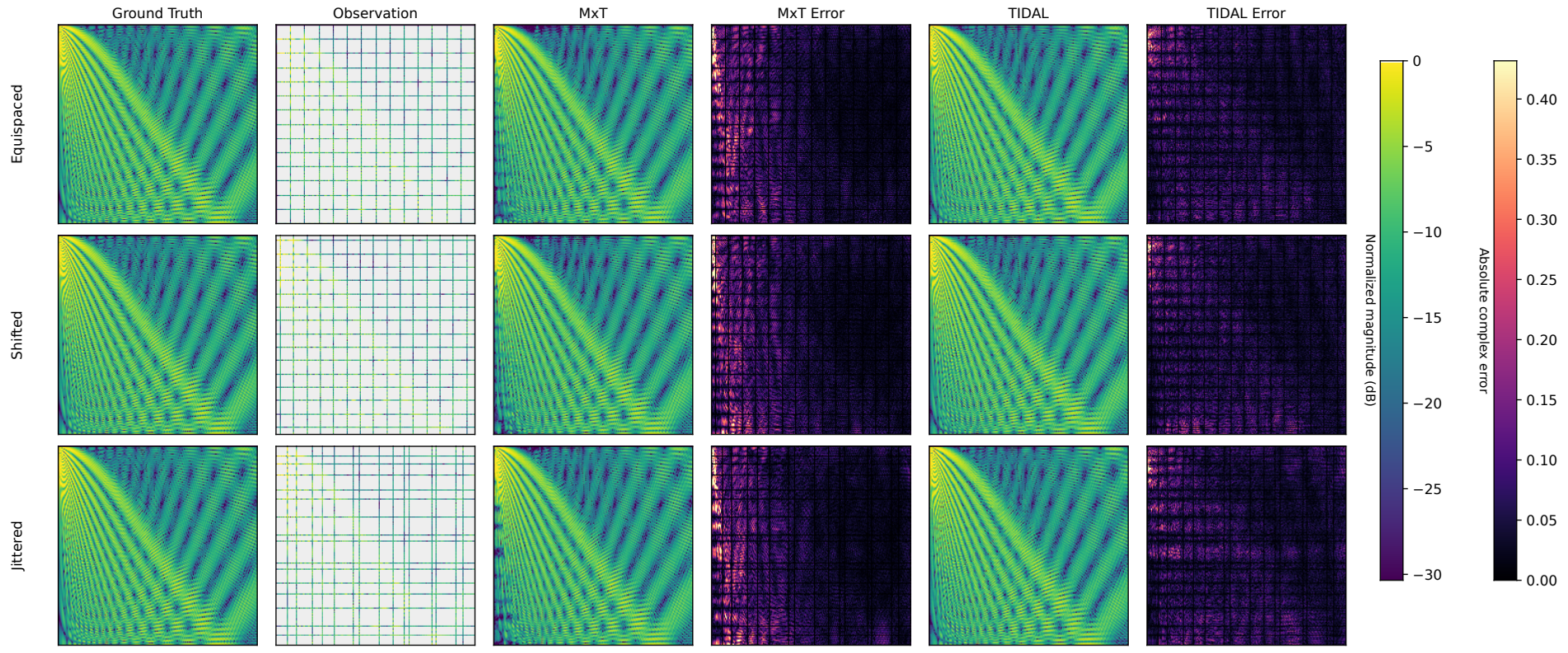}
	\caption{
		Qualitative reconstruction under three sampling geometries.
		Columns show the ground truth, sparse observations, and reconstructions and absolute complex errors of MxT and TIDAL for the same test field and angular slice.
		Field magnitudes and errors use shared scales across all cases.
	}
	\label{fig:qualitative_main}
\end{figure*}

\begin{table*}[t]
	\centering
	\caption{
		Reconstruction performance and inference efficiency under three sampling geometries.
		Results are reported as mean $\pm$ standard deviation over all test fields.
		``Time'' denotes per-field inference time.
		Best results are shown in bold.
	}
	\label{tab:main_results}
	\small
	\setlength{\tabcolsep}{4.0pt}
	\resizebox{\textwidth}{!}{
		\begin{tabular}{llccccccc}
			\toprule
			\multirow{2}{*}{Method}
			& \multirow{2}{*}{Metric}
			& \multicolumn{2}{c}{Equispaced}
			& \multicolumn{2}{c}{Shifted}
			& \multicolumn{2}{c}{Jittered}
			& \multirow{2}{*}{Time (s) $\downarrow$} \\
			\cmidrule(lr){3-4}
			\cmidrule(lr){5-6}
			\cmidrule(lr){7-8}
			&
			& Full & Unobs.
			& Full & Unobs.
			& Full & Unobs.
			& \\
			\midrule
			
			\multirow{2}{*}{LaMO}
			& RMSE $\downarrow$
			& $0.712 \pm 0.068$
			& $0.749 \pm 0.073$
			& $0.704 \pm 0.068$
			& $0.743 \pm 0.072$
			& $0.717 \pm 0.067$
			& $0.757 \pm 0.071$
			& \multirow{2}{*}{$0.91$} \\
			& MAC $\uparrow$
			& $0.488 \pm 0.098$
			& $0.434 \pm 0.110$
			& $0.501 \pm 0.097$
			& $0.444 \pm 0.108$
			& $0.483 \pm 0.096$
			& $0.423 \pm 0.108$
			& \\
			\addlinespace[2pt]
			
			\multirow{2}{*}{SCPNet}
			& RMSE $\downarrow$
			& $0.504 \pm 0.056$
			& $0.541 \pm 0.060$
			& $0.486 \pm 0.055$
			& $0.526 \pm 0.059$
			& $0.542 \pm 0.052$
			& $0.586 \pm 0.056$
			& \multirow{2}{*}{$1.71$} \\
			& MAC $\uparrow$
			& $0.743 \pm 0.056$
			& $0.704 \pm 0.065$
			& $0.761 \pm 0.053$
			& $0.720 \pm 0.062$
			& $0.704 \pm 0.056$
			& $0.653 \pm 0.066$
			& \\
			\addlinespace[2pt]
			
			\multirow{2}{*}{MxT}
			& RMSE $\downarrow$
			& $0.278 \pm 0.073$
			& $0.298 \pm 0.078$
			& $0.281 \pm 0.070$
			& $0.303 \pm 0.076$
			& $0.293 \pm 0.072$
			& $0.317 \pm 0.078$
			& \multirow{2}{*}{$0.72$} \\
			& MAC $\uparrow$
			& $0.918 \pm 0.042$
			& $0.905 \pm 0.049$
			& $0.917 \pm 0.041$
			& $0.903 \pm 0.048$
			& $0.909 \pm 0.044$
			& $0.894 \pm 0.052$
			& \\
			\addlinespace[2pt]
			
			\multirow{2}{*}{DiffusionPDE}
			& RMSE $\downarrow$
			& $0.394 \pm 0.131$
			& $0.423 \pm 0.140$
			& $0.342 \pm 0.137$
			& $0.370 \pm 0.147$
			& $0.331 \pm 0.121$
			& $0.358 \pm 0.131$
			& \multirow{2}{*}{$901.6$} \\
			& MAC $\uparrow$
			& $0.831 \pm 0.108$
			& $0.806 \pm 0.124$
			& $0.867 \pm 0.103$
			& $0.845 \pm 0.118$
			& $0.877 \pm 0.093$
			& $0.857 \pm 0.107$
			& \\
			
			\midrule
			
			\multirow{2}{*}{\textbf{TIDAL}}
			& RMSE $\downarrow$
			& $\mathbf{0.256 \pm 0.083}$
			& $\mathbf{0.269 \pm 0.085}$
			& $\mathbf{0.259 \pm 0.082}$
			& $\mathbf{0.274 \pm 0.085}$
			& $\mathbf{0.278 \pm 0.085}$
			& $\mathbf{0.295 \pm 0.089}$
			& \multirow{2}{*}{$\mathbf{0.14}$} \\
			& MAC $\uparrow$
			& $\mathbf{0.928 \pm 0.047}$
			& $\mathbf{0.920 \pm 0.051}$
			& $\mathbf{0.926 \pm 0.047}$
			& $\mathbf{0.918 \pm 0.052}$
			& $\mathbf{0.916 \pm 0.052}$
			& $\mathbf{0.905 \pm 0.057}$
			& \\
			
			\bottomrule
		\end{tabular}
	}
\end{table*}

\subsection{Main Reconstruction Results}

Table~\ref{tab:main_results} shows that TIDAL achieves the best reconstruction performance among the baselines. 
The larger gains on unobserved locations indicate improved recovery of missing field values rather than reproduction of measured transects.
Although jittered sampling is more challenging than equispaced or shifted layouts, TIDAL maintains its advantage across all geometries. It also achieves the lowest inference time, yielding a favorable accuracy–efficiency trade-off.

Fig.~\ref{fig:qualitative_main} compares TIDAL with MxT, the strongest baseline in Table~\ref{tab:main_results}.
While both recover the dominant propagation pattern, TIDAL better preserves the continuity and the oscillatory interference patterns in unobserved regions, with lower residual errors across all three sampling geometries.

\begin{figure*}[t]
	\centering
	\includegraphics[width=0.85\textwidth]{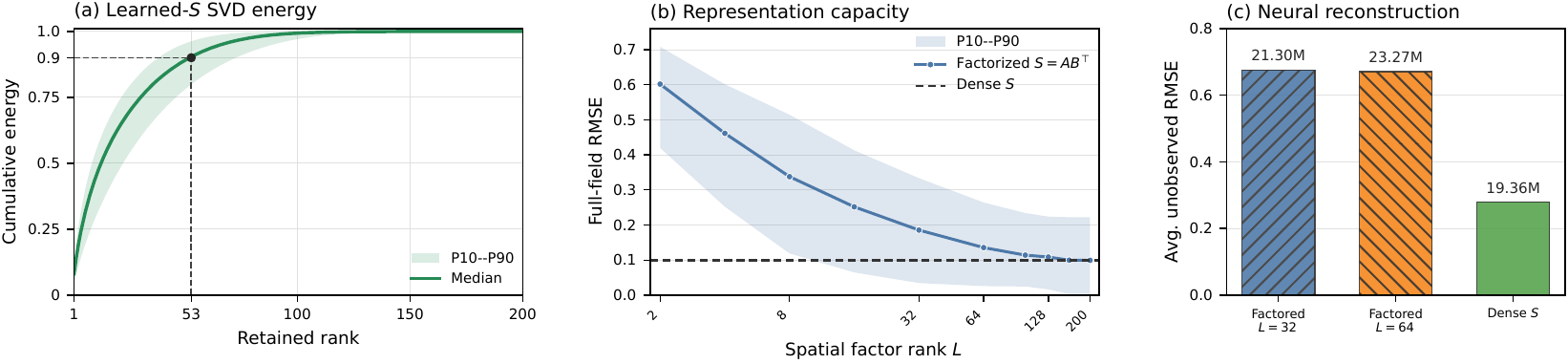}
	\caption{
		Effect of spatial-rank constraints.
		\textbf{(a)} Cumulative singular-value energy of the learned dense spatial components.
		\textbf{(b)} Representation error versus prescribed spatial rank $L$. 
		\textbf{(c)} End-to-end reconstruction with factorized and dense-$S$ decoders. 
		Parameter counts are shown above each bar.
	}
	\label{fig:rank_analysis}
\end{figure*}

\subsection{Effect of Spatial-Rank Constraints}
\label{sec:rank_analysis}

The H/V recoverability analysis in Sec.~\ref{sec:hv_recoverability} considers a low-rank specialization of $S_q$.
TIDAL directly learns dense spatial components within the same $S/c$ representation.
We therefore examine whether low-rank structure emerges in the learned components and the effect of prescribed spatial rank on representation capacity and end-to-end reconstruction.

Fig.~\ref{fig:rank_analysis}(a) shows that the learned dense spatial components exhibit concentrated singular spectra, with the median cumulative energy reaching 90\% at rank 53. 
This indicates that substantial low-rank structure emerges without being explicitly imposed. 
Despite this spectral concentration, Fig.~\ref{fig:rank_analysis}(b) shows that oracle reconstruction error decreases with spatial rank $L$, but a clear gap to dense $S$ remains at moderate ranks and closes only at high ranks. 
Thus, prescribing a moderate spatial rank can limit the representation capacity of the $S/c$ model.

Fig.~\ref{fig:rank_analysis}(c) further examines this effect in end-to-end neural reconstruction.
Increasing the factorized rank from $L=32$ to $L=64$ yields only a marginal improvement, whereas dense spatial components substantially reduce the reconstruction error using fewer trainable parameters than either factorized variant.
Together, these results support the use of data-adaptive spatial components in TIDAL while retaining the physics-derived $S/c$ representation.

\begin{figure*}[t]
	\centering
	\includegraphics[width=0.7\textwidth]{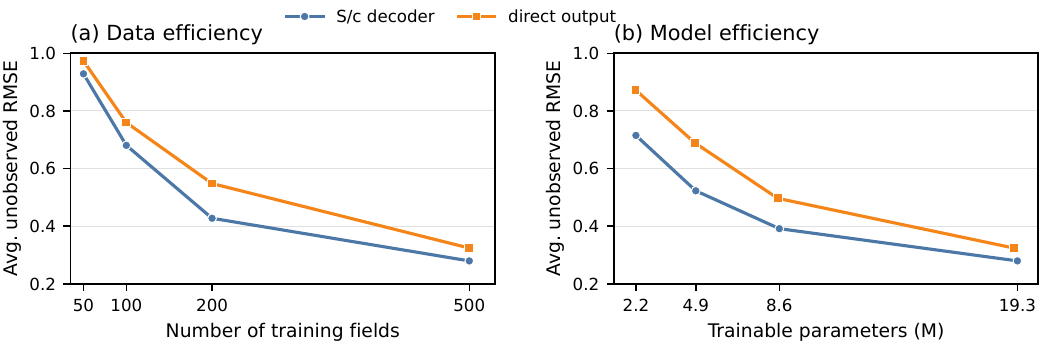}
	\caption{
		Data and model efficiency of $S/c$ decoding.
		Unobserved-region RMSE versus \textbf{(a)} training-set size and \textbf{(b)} model size, comparing S/c and direct-output decoders.
	}
	\label{fig:scaling_analysis}
\end{figure*}

\subsection{Data and Model Efficiency of the $S/c$ Decoder}
\label{sec:efficiency_scaling}

We next examine whether retaining the $S/c$ representation provides a useful inductive bias under different training-set sizes and model capacities.
As shown in Fig.~\ref{fig:scaling_analysis}, the $S/c$ decoder consistently outperforms direct output, with larger gains in limited-data and lightweight-model regimes.
Since paired models have comparable parameter counts, the gains are associated with structured output parameterization rather than model size, particularly under limited supervision or capacity.


\section{Conclusion}

This work demonstrates a representation-level perspective on incorporating physics into neural reconstruction: the mathematical structure of the governing wave equation can directly inform how a physical field is represented. For multi-angle ocean acoustic fields, the resulting $S/c$ representation provides a structured organization of the field and, under a low-rank specialization, connects sparse H/V reconstruction to classical LL1 recoverability theory, enabling sufficient conditions for full-field recovery to be characterized. More broadly, these results illustrate how physics-derived representations can provide a principled interface between wave physics, structured observations, and data-driven reconstruction. Future work will investigate whether this representation-level use of physics can be extended to a broader range of physical fields.

\subsection*{AI Use Statement}

Generative AI tools were used to assist with writing and language polishing.
All AI-assisted edits were reviewed and verified by the authors, who take full responsibility for the final content of this paper.

\bibliography{iclr2027_conference}

@book{stein2011fourier,
	title={Fourier analysis: an introduction},
	author={Stein, Elias M and Shakarchi, Rami},
	volume={1},
	year={2011},
	publisher={Princeton University Press}
}

@article{schmid2021spatial,
	title={Spatial reconstruction of the sound field in a room in the modal frequency range using Bayesian inference},
	author={Schmid, Jonas M and Fernandez-Grande, Efren and Hahmann, Manuel and Gurbuz, Caglar and Eser, Martin and Marburg, Steffen},
	journal={The Journal of the Acoustical Society of America},
	volume={150},
	number={6},
	pages={4385--4394},
	year={2021},
	publisher={AIP Publishing}
}

@article{chen2006finite,
	title={Finite volume coastal ocean},
	author={Chen, Changsheng and Beardsley, Robert C and Cowles, Geoffrey},
	journal={Oceanography},
	volume={19},
	number={1},
	pages={78},
	year={2006},
	publisher={Citeseer}
}

@article{amante2009etopo1,
	title={ETOPO1 arc-minute global relief model: procedures, data sources and analysis},
	author={Amante, Christopher and Eakins, Barry W},
	year={2009}
}

@article{raissi2019physics,
	title={Physics-informed neural networks: A deep learning framework for solving forward and inverse problems involving nonlinear partial differential equations},
	author={Raissi, Maziar and Perdikaris, Paris and Karniadakis, George E},
	journal={Journal of Computational physics},
	volume={378},
	pages={686--707},
	year={2019},
	publisher={Elsevier}
}

@book{plonka2018numerical,
	title     = {Numerical Fourier Analysis},
	author    = {Plonka, Gerlind and Potts, Daniel and Steidl, Gabriele and Tasche, Manfred},
	year      = {2018},
	publisher = {Birkh{\"a}user},
	address   = {Basel}
}

@article{harshman1970foundations,
  title={Foundations of the PARAFAC procedure: Models and conditions for an “explanatory” multi-modal factor analysis},
  author={Harshman, Richard A and others},
  journal={{UCLA} working papers in phonetics},
  volume={16},
  number={1},
  pages={84},
  year={1970},
  publisher={Los Angeles, CA}
}

@article{tucker1966some,
  title={Some mathematical notes on three-mode factor analysis},
  author={Tucker, Ledyard R},
  journal={Psychometrika},
  volume={31},
  number={3},
  pages={279--311},
  year={1966},
  publisher={Springer}
}

@article{sorensen2015coupled,
author = {S{\o}rensen, Mikael and Domanov, Ignat and De Lathauwer, Lieven},
title = {Coupled Canonical Polyadic Decompositions and (Coupled) Decompositions in Multilinear Rank- \$(L\_{r,n},L\_{r,n},1)\$ Terms---Part II: Algorithms},
journal = {SIAM Journal on Matrix Analysis and Applications},
volume = {36},
number = {3},
pages = {1015-1045},
year = {2015},
doi = {10.1137/140956865},

URL = { 
    
        https://doi.org/10.1137/140956865
    
    

},
eprint = { 
    
        https://doi.org/10.1137/140956865
    
    

}
}

@article{candes2010matrix,
  title={Matrix completion with noise},
  author={Candes, Emmanuel J and Plan, Yaniv},
  journal={Proceedings of the IEEE},
  volume={98},
  number={6},
  pages={925--936},
  year={2010},
  publisher={IEEE}
}

@article{ho2019axial,
	title={Axial attention in multidimensional transformers},
	author={Ho, Jonathan and Kalchbrenner, Nal and Weissenborn, Dirk and Salimans, Tim},
	journal={arXiv preprint arXiv:1912.12180},
	year={2019}
}

@article{tancik2020fourier,
	title={Fourier features let networks learn high frequency functions in low dimensional domains},
	author={Tancik, Matthew and Srinivasan, Pratul and Mildenhall, Ben and Fridovich-Keil, Sara and Raghavan, Nithin and Singhal, Utkarsh and Ramamoorthi, Ravi and Barron, Jonathan and Ng, Ren},
	journal={Advances in neural information processing systems},
	volume={33},
	pages={7537--7547},
	year={2020}
}

@inproceedings{
	loshchilov2018decoupled,
	title={Decoupled Weight Decay Regularization},
	author={Ilya Loshchilov and Frank Hutter},
	booktitle={International Conference on Learning Representations},
	year={2019},
	url={https://openreview.net/forum?id=Bkg6RiCqY7},
}

@article{porter1992kraken,
	title={The KRAKEN normal mode program},
	author={Porter, Michael B},
	year={1992}
}

@inproceedings{sun2017underwater,
	title={Underwater acoustic field reconstruction using a compressive sensing approach},
	author={Sun, Jie and Song, Aijun and Yu, Jiancheng and Zhang, Aiqun and Zhang, Fumin},
	booktitle={OCEANS 2017-Anchorage},
	pages={1--5},
	year={2017},
	organization={IEEE}
}

@article{zhu2024strategic,
  title={Strategic deployment in the deep: Principled underwater sensor placement optimization with three-dimensional acoustic map},
  author={Zhu, Xiaohan and Wang, Ye and Fang, Zeyu and Cheng, Lei and Li, Jianlong},
  journal={The Journal of the Acoustical Society of America},
  volume={156},
  number={4},
  pages={2668--2685},
  year={2024},
  publisher={AIP Publishing}
}

@article{sun2020kriged,
	title={A kriged compressive sensing approach to reconstruct acoustic fields from measurements collected by underwater vehicles},
	author={Sun, Jie and Liu, Shijie and Zhang, Fumin and Song, Aijun and Yu, Jiancheng and Zhang, Aiqun},
	journal={IEEE Journal of Oceanic Engineering},
	volume={46},
	number={1},
	pages={294--306},
	year={2020},
	publisher={IEEE}
}

@article{kanatsoulis2018hyperspectral,
	author={Kanatsoulis, Charilaos I. and Fu, Xiao and Sidiropoulos, Nicholas D. and Ma, Wing-Kin},
	journal={IEEE Transactions on Signal Processing}, 
	title={Hyperspectral Super-Resolution: A Coupled Tensor Factorization Approach}, 
	year={2018},
	volume={66},
	number={24},
	pages={6503-6517},
}

@book{canuto2006spectral,
	title={Spectral methods},
	author={Canuto, Claudio and Hussaini, M Youssuff and Quarteroni, Alfio and Zang, Thomas A},
	volume={285},
	year={2006},
	publisher={Springer}
}

@article{ding2021hyperspectral,
	author={Ding, Meng and Fu, Xiao and Huang, Ting-Zhu and Wang, Jun and Zhao, Xi-Le},
	journal={IEEE Journal of Selected Topics in Signal Processing}, 
	title={Hyperspectral Super-Resolution via Interpretable Block-Term Tensor Modeling}, 
	year={2021},
	volume={15},
	number={3},
	pages={641-656},}

@article{kolda2009tensor,
	title={Tensor decompositions and applications},
	author={Kolda, Tamara G and Bader, Brett W},
	journal={SIAM review},
	volume={51},
	number={3},
	pages={455--500},
	year={2009},
	publisher={SIAM}
}

@article{rontogiannis2021block,
	title={Block-term tensor decomposition: Model selection and computation},
	author={Rontogiannis, Athanasios A and Kofidis, Eleftherios and Giampouras, Paris V},
	journal={IEEE Journal of Selected Topics in Signal Processing},
	volume={15},
	number={3},
	pages={464--475},
	year={2021},
	publisher={IEEE}
}

@article{de2008decompositions,
	title={Decompositions of a higher-order tensor in block terms—Part II: Definitions and uniqueness},
	author={De Lathauwer, Lieven},
	journal={SIAM Journal on Matrix Analysis and Applications},
	volume={30},
	number={3},
	pages={1033--1066},
	year={2008},
	publisher={SIAM}
}

@article{wan2009spatial,
	title={Spatial coherence measurements from two L-shape arrays in shallow water},
	author={Wan, Lin and Zhou, Ji-Xun and Rogers, Peter H and Knobles, David P},
	journal={Acoustical Physics},
	volume={55},
	number={3},
	pages={383--392},
	year={2009},
	publisher={Springer}
}

@article{tang2007shallow,
	title={Shallow Water '06: A Joint Acoustic Propagation/Nonlinear Internal Wave Physics Experiment},
	author={Tang, Dajun and Moum, James N and Lynch, James F and Abbot, Phil and Chapman, Ross and Dahl, Peter H and Duda, Timothy F and Gawarkiewicz, Glen and Glenn, Scott and Goff, John A and Graber, Hans and Kemp, John and Maffei, Andrew and Nash, Jonathan D and Newhall, Arthur},
	journal={Oceanography},
	volume={20},
	number={4},
	pages={156--167},
	year={2007}
}

@article{zhang2020spectrum,
	title={Spectrum cartography via coupled block-term tensor decomposition},
	author={Zhang, Guoyong and Fu, Xiao and Wang, Jun and Zhao, Xi-Le and Hong, Mingyi},
	journal={IEEE Transactions on Signal Processing},
	volume={68},
	pages={3660--3675},
	year={2020},
	publisher={IEEE}
}

@InProceedings{Tiwari2025latent,
	title = 	 {Latent Mamba Operator for Partial Differential Equations},
	author =       {Tiwari, Karn and Dutta, Niladri and Krishnan, N M Anoop and Ap, Prathosh},
	booktitle = 	 {Proceedings of the 42nd International Conference on Machine Learning},
	pages = 	 {59639--59667},
	year = 	 {2025},
	volume = 	 {267},
	series = 	 {Proceedings of Machine Learning Research},
	month = 	 {13--19 Jul},
	publisher =    {PMLR}
}

@article{santos2023development,
	title={Development of the senseiver for efficient field reconstruction from sparse observations},
	author={Santos, Javier E and Fox, Zachary R and Mohan, Arvind and O’Malley, Daniel and Viswanathan, Hari and Lubbers, Nicholas},
	journal={Nature Machine Intelligence},
	volume={5},
	number={11},
	pages={1317--1325},
	year={2023},
	publisher={Nature Publishing Group UK London}
}

@article{huang2024diffusionpde,
	title={Diffusionpde: Generative pde-solving under partial observation},
	author={Huang, Jiahe and Yang, Guandao and Wang, Zichen and Park, Jeong Joon},
	journal={Advances in Neural Information Processing Systems},
	volume={37},
	pages={130291--130323},
	year={2024}
}

@inproceedings{ma2025physense,
	author = {Ma, Yuezhou and Wu, Haixu and Zhou, Hang and Weng, Huikun and Wang, Jianmin and Long, Mingsheng},
	booktitle = {Advances in Neural Information Processing Systems},
	doi = {10.52202/085713-1200},
	pages = {35697--35725},
	publisher = {Curran Associates, Inc.},
	title = {PhySense: Sensor Placement Optimization for Accurate Physics Sensing},
	volume = {38, Main Conference},
	year = {2025}
}

@inproceedings{chen2024mamba,
	author    = {Shuang Chen and Amir Atapour-Abarghouei and Haozheng Zhang and Hubert P. H. Shum},
	title     = {MxT: Mamba x Transformer for Image Inpainting},
	booktitle = {35th British Machine Vision Conference 2024, {BMVC} 2024, Glasgow, UK, November 25-28, 2024},
	publisher = {BMVA},
	year      = {2024},
	url       = {https://papers.bmvc2024.org/0295.pdf}
}

@inproceedings{xia2023scpnet,
	title={Scpnet: Semantic scene completion on point cloud},
	author={Xia, Zhaoyang and Liu, Youquan and Li, Xin and Zhu, Xinge and Ma, Yuexin and Li, Yikang and Hou, Yuenan and Qiao, Yu},
	booktitle={2023 IEEE/CVF Conference on Computer Vision and Pattern Recognition (CVPR)},
	pages={17642--17651},
	year={2023},
	organization={IEEE}
}

@book{jensen2011computational,
	title={Computational ocean acoustics},
	author={Jensen, Finn B and Kuperman, William A and Porter, Michael B and Schmidt, Henrik and Tolstoy, Alexandra},
	volume={2011},
	year={2011},
	publisher={Springer}
}

@article{chen2025frequency,
	title={Frequency-domain physics-informed neural network for accurate reconstruction of 3D acoustic fields under sparse and multi-frequency measurements},
	author={Chen, Fangchao and Xiao, Youhong and Yu, Liang and Jiang, Laixu},
	journal={Neural Networks},
	pages={108476},
	year={2025},
	publisher={Elsevier}
}

@article{chen2026physics,
	title={Physics-aware graph learning for sound-field reconstruction from sparse measurements},
	author={Chen, Fangchao and Xiao, Youhong and Yu, Liang and Jiang, Laixu},
	journal={The Journal of the Acoustical Society of America},
	volume={159},
	number={3},
	pages={1973--1986},
	year={2026},
	publisher={AIP Publishing}
}

@inproceedings{damiano2025zero,
	title={A zero-shot physics-informed dictionary learning approach for sound field reconstruction},
	author={Damiano, Stefano and Miotello, Federico and Pezzoli, Mirco and Bernardini, Alberto and Antonacci, Fabio and Sarti, Augusto and Van Waterschoot, Toon},
	booktitle={ICASSP 2025-2025 IEEE International Conference on Acoustics, Speech and Signal Processing (ICASSP)},
	pages={1--5},
	year={2025},
	organization={IEEE}
}

@article{zhao2024recfno,
	title={RecFNO: A resolution-invariant flow and heat field reconstruction method from sparse observations via Fourier neural operator},
	author={Zhao, Xiaoyu and Chen, Xiaoqian and Gong, Zhiqiang and Zhou, Weien and Yao, Wen and Zhang, Yunyang},
	journal={International Journal of Thermal Sciences},
	volume={195},
	pages={108619},
	year={2024},
	publisher={Elsevier}
}
\bibliographystyle{iclr2027_conference}

\appendix

\section{Theory Details for the $S/c$ Representation and Complex H/V Recoverability}
\label{app:identifiability}

This appendix reviews additional derivation details for the development in Sec.~\ref{sec:theory}.
We first introduce the preliminaries of the normal mode separation and finite-dimensional angular representation underlying Proposition~\ref{prop:finite_sc_representation}, including the modal and angular truncation errors.
We then summarize the classical complex LL1 essential-uniqueness result \citep{de2008decompositions} used for the low-rank specialization and provide the complete symmetric conditions associated with the coupled slab-sampling result of \citet{zhang2020spectrum} for structured H/V observations.

\subsection{Details of the Finite-Dimensional $S/c$ Representation}
\label{app:angular_approximation}

We first provide the normal-mode and angular approximation details underlying Proposition~\ref{prop:finite_sc_representation}.
For a fixed direction $\alpha$, substituting the separated form
$p_{\alpha}(z,r)=\psi(z)\phi(r)$
into \eqref{eq:helmholtz} yields \citep{jensen2011computational}
\begin{equation}
	\frac{\mathrm{d}^2\psi_{\ell,\alpha}}{\mathrm{d}z^2}
	+
	\left[
	\frac{\omega^2}{c_{\mathrm{s},\alpha}^{2}(z)}
	-
	k_{r,\ell,\alpha}^{2}
	\right]
	\psi_{\ell,\alpha}
	=
	0,
	\label{eq:app_modal_depth}
\end{equation}
and
\begin{equation}
	\frac{\mathrm{d}^2\phi_{\ell,\alpha}}{\mathrm{d}r^2}
	+
	\frac{1}{r}
	\frac{\mathrm{d}\phi_{\ell,\alpha}}{\mathrm{d}r}
	+
	k_{r,\ell,\alpha}^{2}
	\phi_{\ell,\alpha}
	=
	0,
	\label{eq:app_modal_range}
\end{equation}
where $k_{r,\ell,\alpha}$ is the horizontal wavenumber associated with the $\ell$th mode.
Together with the prescribed surface and bottom boundary conditions, \eqref{eq:app_modal_depth} defines the depth eigenvalue problem, while \eqref{eq:app_modal_range} gives the corresponding range-dependent responses for each eigenmode \citep{jensen2011computational}. The resulting eigenfunctions form the normal-mode basis, and linear superposition gives the modal expansion
\begin{equation}
	p(z,r,\alpha)
	=
	\sum_{\ell\geq 1}
	\psi_{\ell,\alpha}(z)\phi_{\ell,\alpha}(r).
\end{equation}
For a finite modal approximation, retaining \(L_m\) modes gives
\begin{equation}
	p^{(L_m)}(z,r,\alpha)
	=
	\sum_{\ell=1}^{L_m}
	\psi_{\ell,\alpha}(z)
	\phi_{\ell,\alpha}(r),
	\label{eq:app_modal_truncation}
\end{equation}
with the discarded modal contribution
$p(z,r,\alpha)-p^{(L_m)}(z,r,\alpha)$.

We next apply the angular expansion to $p^{(L_m)}$ \citep{stein2011fourier}.
Its Fourier coefficients are
\begin{equation}
	\begin{aligned}
		\widehat{p}^{(L_m)}_{n}(z,r)
		&=
		\frac{1}{2\pi}
		\int_{0}^{2\pi}
		p^{(L_m)}(z,r,\alpha)
		e^{-\mathrm{i}n\alpha}\,\mathrm{d}\alpha
		\\
		&=
		\frac{1}{2\pi}
		\sum_{\ell=1}^{L_m}
		\int_{0}^{2\pi}
		\psi_{\ell,\alpha}(z)
		\phi_{\ell,\alpha}(r)
		e^{-\mathrm{i}n\alpha}\,\mathrm{d}\alpha .
	\end{aligned}
	\label{eq:app_modal_fourier_coeff}
\end{equation}
Thus, each angular coefficient collects the directional variation of the retained modal contributions into a depth--range field.
The superscript $(L_m)$ is suppressed in Sec.~\ref{sec:physics_btd} for notational simplicity.

Retaining the angular orders $|n|\leq N_{\alpha}$ gives
\begin{equation}
	p^{(L_m,N_{\alpha})}(z,r,\alpha)
	=
	\sum_{|n|\leq N_{\alpha}}
	\widehat{p}^{(L_m)}_{n}(z,r)
	e^{\mathrm{i}n\alpha}.
	\label{eq:app_angular_truncation}
\end{equation}
With \(Q=2N_\alpha+1\), let \(n_q=-N_\alpha+q-1\) and define
\begin{equation}
	S_q(z,r)\triangleq \widehat p^{(L_m)}_{n_q}(z,r),
	\qquad
	c_q(\alpha)\triangleq e^{\mathrm{i}n_q\alpha},
	\qquad q=1,\ldots,Q.
\end{equation}
\eqref{eq:app_angular_truncation} can be then written as
\begin{equation}
	p^{(L_m,N_\alpha)}(z,r,\alpha)
	=
	\sum_{q=1}^{Q}S_q(z,r)c_q(\alpha).
	\label{eq:app_continuous_sc}
\end{equation}
Thus, \(S_q\) is the depth--range coefficient field associated with the retained angular Fourier harmonic \(c_q\).
%
Hence, the S/c form directly expresses the truncated angular Fourier expansion through paired depth--range coefficient fields and angular harmonics. 

Sampling \eqref{eq:app_continuous_sc} at the depth, range, and angular grids defined in Sec.~\ref{sec:problem_formulation}, with
\begin{equation}
	[S_q]_{ij}=S_q(z_i,r_j),
	\qquad
	[c_q]_k=c_q(\alpha_k),
\end{equation}
gives
\begin{equation}
	\mathcal{X}^{(L_m,N_{\alpha})}
	=
	\sum_{q=1}^{Q}
	S_q\circ c_q.
	\label{eq:app_sampled_sc}
\end{equation}
Here, \(Q\) is the number of retained angular Fourier harmonics, whereas \(K\) denotes the number of sampled angular directions.

The residual in Proposition~\ref{prop:finite_sc_representation} follows from the two successive truncations,
\begin{equation}
	\begin{aligned}
		p(z,r,\alpha)
		-
		p^{(L_m,N_{\alpha})}(z,r,\alpha)
		&=
		\left[
		p(z,r,\alpha)
		-
		p^{(L_m)}(z,r,\alpha)
		\right]
		\\
		&\quad+
		\left[
		p^{(L_m)}(z,r,\alpha)
		-
		p^{(L_m,N_{\alpha})}(z,r,\alpha)
		\right].
	\end{aligned}
	\label{eq:app_error_decomposition}
\end{equation}
The first term is the discarded modal contribution.
For the second term, Parseval's identity \citep{canuto2006spectral} gives, for each fixed $(z,r)$,
\begin{equation}
	\frac{1}{2\pi}
	\int_{0}^{2\pi}
	\left|
	p^{(L_m)}(z,r,\alpha)
	-
	p^{(L_m,N_{\alpha})}(z,r,\alpha)
	\right|^{2}
	\,\mathrm{d}\alpha
	=
	\sum_{|n|>N_{\alpha}}
	\left|
	\widehat{p}^{(L_m)}_{n}(z,r)
	\right|^{2}.
	\label{eq:app_fourier_tail}
\end{equation}
If the angular dependence has $s$ square-integrable derivatives, standard Fourier approximation further yields
\begin{equation}
	\begin{aligned}
		&
		\left(
		\frac{1}{2\pi}
		\int_{0}^{2\pi}
		\left|
		p^{(L_m)}(z,r,\alpha)
		-
		p^{(L_m,N_{\alpha})}(z,r,\alpha)
		\right|^{2}
		\,\mathrm{d}\alpha
		\right)^{1/2}
		\\
		&\qquad\leq
		(N_{\alpha}+1)^{-s}
		\left(
		\frac{1}{2\pi}
		\int_{0}^{2\pi}
		\left|
		\partial_{\alpha}^{s}
		p^{(L_m)}(z,r,\alpha)
		\right|^{2}
		\,\mathrm{d}\alpha
		\right)^{1/2},
	\end{aligned}
	\label{eq:app_fourier_error_bound}
\end{equation}
which gives the $\mathcal{O}(N_{\alpha}^{-s})$ angular truncation rate stated in Proposition~\ref{prop:finite_sc_representation} \citep{canuto2006spectral,plonka2018numerical}.
Sampling \eqref{eq:app_error_decomposition} on the spatial--angular grid produces the residual tensor $\mathcal{E}$ in \eqref{eq:sc_representation}.

\subsection{Classical LL1 Results for the Low-Rank $S/c$ Specialization}
\label{app:ll1_preliminaries}

We next summarize the classical LL1 essential-uniqueness result used in the complex-valued recoverability analysis.
Consider an LL1 model~\citep{de2008decompositions}
\begin{equation}
	\mathcal{X}
	=
	\sum_{q=1}^{Q}
	\left(A_q B_q^{\top}\right)\circ c_q,
	\qquad
	A_q\in\mathbb{C}^{I\times L},
	\quad
	B_q\in\mathbb{C}^{J\times L},
	\quad
	c_q\in\mathbb{C}^{K}.
	\label{eq:app_ll1_model}
\end{equation}
The corresponding spatial block is
\begin{equation}
	S_q=A_qB_q^{\top}\in\mathbb{C}^{I\times J}.
\end{equation}

Following \citet{de2008decompositions}, define the partitioned factor matrices
\begin{equation}
	A=[A_1,\ldots,A_Q],
	\qquad
	B=[B_1,\ldots,B_Q],
	\qquad
	C=[c_1,\ldots,c_Q].
\end{equation}
For the partitioned matrix $A$, the block $k$-rank $k'_A$ is the largest integer $r$ such that every collection of $r$ blocks contains $rL$ linearly independent columns; $k'_B$ is defined analogously.
The ordinary Kruskal rank of $C$ is denoted by $k_C$.

We assume that the real and imaginary parts of all entries of $(A,B,C)$ are jointly absolutely continuous with respect to Lebesgue measure on the corresponding real Euclidean space.
Under this standard genericity assumption, the relevant block and Kruskal ranks attain their maximal values almost surely \citep{de2008decompositions}.
In particular,
\begin{equation}
	k'_A
	=
	\min\left(
	\left\lfloor\frac{I}{L}\right\rfloor,
	Q
	\right),
	\qquad
	k'_B
	=
	\min\left(
	\left\lfloor\frac{J}{L}\right\rfloor,
	Q
	\right),
	\label{eq:app_generic_block_ranks}
\end{equation}
and, when $K\geq Q$,
\begin{equation}
	k_C=Q.
	\label{eq:app_generic_kc}
\end{equation}
Moreover, for $I\geq L$ and $J\geq L$, $A_q$ and $B_q$ have full column rank almost surely, and hence
\begin{equation}
	\operatorname{rank}(S_q)
	=
	\operatorname{rank}(A_qB_q^{\top})
	=
	L
\end{equation}
for every $q$.

The classical LL1 essential-uniqueness result applies over both $\mathbb{R}$ and $\mathbb{C}$ \citep[Theorem~4.4]{de2008decompositions}.
For \eqref{eq:app_ll1_model}, a sufficient condition is
\begin{equation}
	k_C=Q,
	\qquad
	k'_A+k'_B\geq Q+2.
	\label{eq:app_ll1_uniqueness_condition}
\end{equation}
Therefore, for generic factors over $\mathbb{C}$, with
\begin{equation}
	I\geq L,
	\qquad
	J\geq L,
	\qquad
	K\geq Q,
	\label{eq:app_ll1_dimension_conditions}
\end{equation}
the condition
\begin{equation}
	\min\left(
	\left\lfloor\frac{I}{L}\right\rfloor,
	Q
	\right)
	+
	\min\left(
	\left\lfloor\frac{J}{L}\right\rfloor,
	Q
	\right)
	\geq Q+2
	\label{eq:app_ll1_generic_condition}
\end{equation}
is sufficient for essential uniqueness of the LL1 decomposition.

Under these conditions, the block terms $\{S_q,c_q\}_{q=1}^{Q}$ are identifiable up to a common permutation and reciprocal nonzero complex scaling.
Equivalently, if another $Q$-term LL1 decomposition represents the same tensor, then there exist a permutation $\pi$ of $\{1,\ldots,Q\}$ and nonzero complex scalars $\alpha_1,\ldots,\alpha_Q$ such that
\begin{equation}
	\widetilde{S}_q
	=
	\alpha_q S_{\pi(q)},
	\qquad
	\widetilde{c}_q
	=
	\alpha_q^{-1}c_{\pi(q)}.
	\label{eq:app_ll1_ambiguity}
\end{equation}

The individual factors $A_q$ and $B_q$ are not uniquely determined.
Indeed, for any nonsingular
$F_q\in\mathbb{C}^{L\times L}$,
\begin{equation}
	A_qB_q^{\top}
	=
	(A_qF_q)
	(B_qF_q^{-\top})^{\top}.
\end{equation}
Thus, the identifiable spatial quantity is the block $S_q=A_qB_q^{\top}$ rather than its particular matrix factorization.

\subsection{H/V Recoverability from Coupled Slab Sampling}
\label{app:hv_recoverability}

We provide the complete specialization of \citet[Theorem~3]{zhang2020spectrum} to the H/V observation model in Sec.~\ref{sec:hv_recoverability}.
Related recoverability results for coupled tensor models under structured observations have also been developed for hyperspectral super-resolution \citep{kanatsoulis2018hyperspectral,ding2021hyperspectral}.

Let $P_H\in\{0,1\}^{I_H\times I}$ and $Q_V\in\{0,1\}^{J_V\times J}$ denote the coordinate-selection matrices associated with the selected horizontal and vertical transects, respectively, where $I_H=|\mathcal{I}_H|$ and $J_V=|\mathcal{J}_V|$.
For the exact low-rank model in \eqref{eq:low_rank_sc_model}, the corresponding observations are
\begin{align}
	\mathcal{Y}^{H}
	&=
	\sum_{q=1}^{Q}
	\left((P_HA_q)B_q^{\top}\right)\circ c_q
	=
	\sum_{q=1}^{Q}
	(P_HS_q)\circ c_q,
	\label{eq:app_h_observation}
	\\
	\mathcal{Y}^{V}
	&=
	\sum_{q=1}^{Q}
	\left(A_q(Q_VB_q)^{\top}\right)\circ c_q
	=
	\sum_{q=1}^{Q}
	(S_qQ_V^{\top})\circ c_q.
	\label{eq:app_v_observation}
\end{align}
Coordinate selection therefore preserves the LL1 form, and $\mathcal{Y}^{H}$ and $\mathcal{Y}^{V}$ form two coupled LL1 tensors with shared angular factors $\{c_q\}_{q=1}^{Q}$.
Accordingly, the H/V observation model is a special case of the coupled slab-sampling framework analyzed by \citet{zhang2020spectrum}.

In \citet[Theorem~3]{zhang2020spectrum}, a special case is the present H/V observation model, where the two coupled tensors contain $Q$ block terms and retain all $K$ samples along the shared angular mode.
The corresponding third-mode overlap condition therefore reduces to
\begin{equation}
	K\geq Q.
	\label{eq:app_hv_angular_condition}
\end{equation}

The remaining spatial sampling conditions yield two symmetric cases.
The first requires
\begin{equation}
	I_H\geq 2L,
	\qquad
	J\geq LQ,
	\label{eq:app_h_condition_dims}
\end{equation}
together with
\begin{equation}
	\min\left(
	\left\lfloor\frac{I_H}{L}\right\rfloor,
	Q
	\right)
	+
	\min\left(
	\left\lfloor\frac{J}{L}\right\rfloor,
	Q
	\right)
	\geq Q+2.
	\label{eq:app_h_condition}
\end{equation}
Alternatively, the symmetric condition requires
\begin{equation}
	J_V\geq 2L,
	\qquad
	I\geq LQ,
	\label{eq:app_v_condition_dims}
\end{equation}
together with
\begin{equation}
	\min\left(
	\left\lfloor\frac{J_V}{L}\right\rfloor,
	Q
	\right)
	+
	\min\left(
	\left\lfloor\frac{I}{L}\right\rfloor,
	Q
	\right)
	\geq Q+2.
	\label{eq:app_v_condition}
\end{equation}
Together with \eqref{eq:app_hv_angular_condition}, either set of conditions is sufficient for recoverability under the coupled slab-sampling model.

For complex-valued factors, genericity is understood with respect to the joint real and imaginary parts of the factors.
Under this assumption, the LL1 essential-uniqueness condition used in the slab-sampling argument remains valid over $\mathbb{C}$ by \citet[Theorem~4.4]{de2008decompositions}, as summarized in Appendix~\ref{app:ll1_preliminaries}.
The subsequent coupling argument relies on full-rank linear relations and the alignment of the shared angular factors, which are unchanged over $\mathbb{C}$.

Under the above conditions, the coupled observations $(\mathcal{Y}^{H},\mathcal{Y}^{V})$ identify the block terms $\{S_q,c_q\}_{q=1}^{Q}$ up to a common permutation and reciprocal nonzero complex scaling.
These ambiguities do not affect the recomposed tensor because
\begin{equation}
	(\alpha_qS_q)\circ(\alpha_q^{-1}c_q)
	=
	S_q\circ c_q.
\end{equation}
Consequently,
\begin{equation}
	\mathcal{X}
	=
	\sum_{q=1}^{Q}S_q\circ c_q
\end{equation}
is uniquely determined by the structured H/V observations.

\section{Implementation Details}
\label{app:implementation}

\subsection{Data Processing and Observation Masks}
\label{app:data_processing_and_mask}

The dataset spans acoustic frequencies of 100, 150, and 200~Hz, source depths of 30, 50, 80, and 120~m, and three source locations, with approximately balanced coverage across these settings.
For preprocessing, each field is treated independently by compensating normal-mode geometric spreading along range and scaling the resulting complex field by its 99.5th-percentile magnitude.  Real and imaginary parts are represented as two channels. During training, the numbers of horizontal and vertical transects are sampled independently from $\{10,15,20\}$, and masks are resampled at every epoch from equispaced, shifted, and jittered sampling geometries.

\paragraph{H/V sampling geometries.}
Let $\mathcal{I}_H\subseteq[I]$ and $\mathcal{J}_V\subseteq[J]$ denote the selected horizontal and vertical transects, respectively, with
\begin{equation}
	\Omega_{\mathrm{HV}}
	=
	\big(
	(\mathcal{I}_H\times[J])
	\cup
	([I]\times\mathcal{J}_V)
	\big)\times[K].
\end{equation}
The same spatial observation pattern is shared across all $K$ angular slices.
The sets $\mathcal{I}_H$ and $\mathcal{J}_V$ are generated independently using the same sampling rule along the depth and range axes.

For an axis of length $n$ with $s$ selected transects, using zero-based index
$m=0,\ldots,s-1$, we consider three sampling geometries.

\emph{Equispaced.}
The selected indices are uniformly distributed between the two endpoints:
\begin{equation}
	\left\{
	\operatorname{round}\!\left(
	\frac{m(n-1)}{s-1}
	\right)
	\right\}_{m=0}^{s-1}.
\end{equation}

\emph{Shifted.}
An equally spaced grid is shifted by a random phase:
\begin{equation}
	\left\{
	\left\lfloor
	\frac{(m+\phi)n}{s}
	\right\rfloor
	\right\}_{m=0}^{s-1},
	\qquad
	\phi\sim\mathcal{U}[0,1).
\end{equation}
The phase shifts are sampled independently for the depth and range axes.

\emph{Jittered.}
The axis is partitioned into $s$ approximately equal and non-overlapping intervals, and one transect is sampled uniformly from each interval:
\begin{equation}
	\xi_m
	\sim
	\operatorname{Unif}
	\left(
	\left\{
	\left\lfloor\frac{mn}{s}\right\rfloor,
	\ldots,
	\left\lfloor\frac{(m+1)n}{s}\right\rfloor-1
	\right\}
	\right),
	\qquad
	m=0,\ldots,s-1.
\end{equation}
The jittered selections are sampled independently along the depth and range axes.

\subsection{Model and Objective}

TIDAL uses $Q_N=40$ components, latent width 384, and eight axial-attention blocks with eight heads each.  Its input comprises
the 72-angle real and imaginary observed fields, a shared spatial observation mask, and normalized depth and range coordinates (147 channels in total).
The angular branch uses the normalized angle coordinate together with eight sine/cosine Fourier frequencies.

We optimize the sum of relative complex mean-squared-error terms over the full field, observed locations, unobserved locations, and the far half of the range axis, with weights $1$, $1$, $1$, and $0.5$, respectively.  We additionally apply a finite-difference smoothness penalty to the learned spatial components with weight $10^{-4}$.

\subsection{Optimization and Model Selection}

Models are trained for 150 epochs with batch size one using AdamW \citep{loshchilov2018decoupled}, an initial learning rate of $3\times10^{-4}$, and weight decay $10^{-5}$.  We use a StepLR scheduler that decays the learning rate by a factor of 0.5 every 40 epochs.  The random seed is fixed to zero.  We select the checkpoint with the
lowest validation full-field relative RMSE and use this checkpoint for all test
evaluations.

\subsection{Evaluation Protocol}

Unless stated otherwise, evaluation uses 15 horizontal and 15 vertical transects under the equispaced, shifted, and jittered sampling geometries. 
We report full-field and unobserved-region relative RMSE and MAC.
Let $\mathbf{x}$ and $\hat{\mathbf{x}}$ denote the vectorized reference and reconstructed fields, respectively.
MAC is defined as
\begin{equation}
	\mathrm{MAC}(\mathbf{x},\hat{\mathbf{x}})
	=
	\frac{\left|\mathbf{x}^{H}\hat{\mathbf{x}}\right|^2}
	{\left(\mathbf{x}^{H}\mathbf{x}\right)
		\left(\hat{\mathbf{x}}^{H}\hat{\mathbf{x}}\right)}.
\end{equation}
Both metrics are computed jointly over all 72 angles for each test field, and the mean and standard deviation are then reported over the 80 test fields.
All methods are evaluated on the same held-out acoustic fields and observation masks. Inference time is measured on a single NVIDIA A800 GPU.

\subsection{Baseline Implementation and Adaptation}
\label{app:baseline_implementation}

All learning-based baselines are trained on the same 500 acoustic training fields and use the same field normalization, train/validation/test split, and H/V sampling protocol as TIDAL. We retain the main backbone of each baseline while adapting its input, output, and observation-conditioning interface to complex acoustic-field reconstruction.

MxT is adapted from image inpainting to complex field completion on individual depth--range slices. The real and imaginary parts of the sparse observations are provided together with the binary observation mask, and the output head predicts the corresponding dense complex field. The original Mamba--Transformer generator is retained.

LaMO is implemented as a regular-grid operator acting on depth--range slices. Each spatial location is represented by its real and imaginary observations together with the sampling mask, while normalized depth and range coordinates are provided as positional inputs. The model directly predicts the dense real and imaginary field components.

We adapt the completion backbone of SCPNet from semantic scene completion to continuous complex-field reconstruction. The semantic-specific prediction heads are removed, while the multi-scale 3D completion architecture is retained. The network takes the real and imaginary observations together with the occupancy mask and predicts the dense complex field over depth, range, and angle.

DiffusionPDE is adapted to model the real and imaginary components of individual depth--range acoustic slices using its original diffusion-based generative formulation. Sparse measurements are incorporated during iterative sampling through observation-gradient guidance rather than being directly concatenated with the diffusion-model input. The PDE-guidance term is disabled because the physical parameters required to instantiate the PDE constraint are unavailable in our reconstruction setting; reconstruction is therefore conditioned only on the observed acoustic measurements.

\section{Additional Results under Different H/V Sampling Budgets}

To examine whether the main results depend on the default H15/V15 sampling budget, we further evaluate all methods under H10/V10 and H20/V20 observations.
The training protocol is unchanged, and all models are evaluated on the same 80 test fields and observation masks under the equispaced, shifted, and jittered sampling geometries.
Table~\ref{tab:sampling_budget} reports RMSE and MAC over the full field and unobserved spatial locations.

\begin{table*}[t]
	\centering
	\caption{
		Reconstruction performance under H10/V10 and H20/V20 sampling.
		Results are reported as mean $\pm$ standard deviation over 80 test fields.
		Best results are shown in bold.
	}
	\label{tab:sampling_budget}
	\small
	\setlength{\tabcolsep}{3.8pt}
	\resizebox{\textwidth}{!}{
		\begin{tabular}{lllcccccc}
			\toprule
			\multirow{2}{*}{Budget}
			& \multirow{2}{*}{Method}
			& \multirow{2}{*}{Metric}
			& \multicolumn{2}{c}{Equispaced}
			& \multicolumn{2}{c}{Shifted}
			& \multicolumn{2}{c}{Jittered} \\
			\cmidrule(lr){4-5}
			\cmidrule(lr){6-7}
			\cmidrule(lr){8-9}
			&
			&
			& Full & Unobs.
			& Full & Unobs.
			& Full & Unobs. \\
			\midrule
			
			\multirow{10}{*}{H10/V10}
			
			& \multirow{2}{*}{LaMO}
			& RMSE $\downarrow$
			& $0.792 \pm 0.060$
			& $0.820 \pm 0.063$
			& $0.770 \pm 0.067$
			& $0.800 \pm 0.069$
			& $0.781 \pm 0.061$
			& $0.812 \pm 0.064$ \\
			&
			& MAC $\uparrow$
			& $0.370 \pm 0.096$
			& $0.325 \pm 0.104$
			& $0.404 \pm 0.103$
			& $0.356 \pm 0.111$
			& $0.387 \pm 0.096$
			& $0.337 \pm 0.104$ \\
			\addlinespace[2pt]
			
			& \multirow{2}{*}{SCPNet}
			& RMSE $\downarrow$
			& $0.693 \pm 0.044$
			& $0.724 \pm 0.046$
			& $0.659 \pm 0.047$
			& $0.694 \pm 0.049$
			& $0.676 \pm 0.045$
			& $0.712 \pm 0.047$ \\
			&
			& MAC $\uparrow$
			& $0.518 \pm 0.061$
			& $0.474 \pm 0.067$
			& $0.564 \pm 0.062$
			& $0.517 \pm 0.068$
			& $0.541 \pm 0.060$
			& $0.491 \pm 0.066$ \\
			\addlinespace[2pt]
			
			& \multirow{2}{*}{MxT}
			& RMSE $\downarrow$
			& $0.329 \pm 0.082$
			& $0.343 \pm 0.086$
			& $0.329 \pm 0.079$
			& $0.346 \pm 0.084$
			& $0.343 \pm 0.080$
			& $0.362 \pm 0.084$ \\
			&
			& MAC $\uparrow$
			& $0.886 \pm 0.056$
			& $0.875 \pm 0.061$
			& $0.886 \pm 0.054$
			& $0.874 \pm 0.060$
			& $\mathbf{0.876 \pm 0.057}$
			& $0.863 \pm 0.063$ \\
			\addlinespace[2pt]
			
			& \multirow{2}{*}{DiffusionPDE}
			& RMSE $\downarrow$
			& $0.622 \pm 0.135$
			& $0.651 \pm 0.141$
			& $0.535 \pm 0.132$
			& $0.563 \pm 0.139$
			& $0.538 \pm 0.132$
			& $0.566 \pm 0.138$ \\
			&
			& MAC $\uparrow$
			& $0.605 \pm 0.157$
			& $0.570 \pm 0.170$
			& $0.701 \pm 0.139$
			& $0.670 \pm 0.152$
			& $0.699 \pm 0.137$
			& $0.667 \pm 0.151$ \\
			\addlinespace[2pt]
			
			& \multirow{2}{*}{\textbf{TIDAL}}
			& RMSE $\downarrow$
			& $\mathbf{0.317 \pm 0.095}$
			& $\mathbf{0.329 \pm 0.097}$
			& $\mathbf{0.311 \pm 0.091}$
			& $\mathbf{0.324 \pm 0.095}$
			& $\mathbf{0.343 \pm 0.093}$
			& $\mathbf{0.358 \pm 0.097}$ \\
			&
			& MAC $\uparrow$
			& $\mathbf{0.891 \pm 0.065}$
			& $\mathbf{0.883 \pm 0.069}$
			& $\mathbf{0.895 \pm 0.061}$
			& $\mathbf{0.887 \pm 0.066}$
			& $0.874 \pm 0.068$
			& $\mathbf{0.863 \pm 0.074}$ \\
			
			\midrule
			
			\multirow{10}{*}{H20/V20}
			
			& \multirow{2}{*}{LaMO}
			& RMSE $\downarrow$
			& $0.671 \pm 0.064$
			& $0.716 \pm 0.071$
			& $0.667 \pm 0.064$
			& $0.714 \pm 0.070$
			& $0.678 \pm 0.066$
			& $0.725 \pm 0.072$ \\
			&
			& MAC $\uparrow$
			& $0.547 \pm 0.088$
			& $0.484 \pm 0.102$
			& $0.552 \pm 0.087$
			& $0.487 \pm 0.101$
			& $0.538 \pm 0.091$
			& $0.470 \pm 0.105$ \\
			\addlinespace[2pt]
			
			& \multirow{2}{*}{SCPNet}
			& RMSE $\downarrow$
			& $0.409 \pm 0.052$
			& $0.451 \pm 0.058$
			& $0.409 \pm 0.053$
			& $0.455 \pm 0.058$
			& $0.450 \pm 0.054$
			& $0.501 \pm 0.059$ \\
			&
			& MAC $\uparrow$
			& $0.830 \pm 0.043$
			& $0.793 \pm 0.053$
			& $0.830 \pm 0.043$
			& $0.790 \pm 0.054$
			& $0.795 \pm 0.049$
			& $0.746 \pm 0.060$ \\
			\addlinespace[2pt]
			
			& \multirow{2}{*}{MxT}
			& RMSE $\downarrow$
			& $0.250 \pm 0.068$
			& $0.276 \pm 0.075$
			& $0.253 \pm 0.065$
			& $0.281 \pm 0.072$
			& $0.263 \pm 0.068$
			& $0.292 \pm 0.075$ \\
			&
			& MAC $\uparrow$
			& $0.933 \pm 0.036$
			& $0.919 \pm 0.043$
			& $0.932 \pm 0.035$
			& $0.916 \pm 0.043$
			& $0.927 \pm 0.037$
			& $0.910 \pm 0.046$ \\
			\addlinespace[2pt]
			
			& \multirow{2}{*}{DiffusionPDE}
			& RMSE $\downarrow$
			& $0.321 \pm 0.133$
			& $0.354 \pm 0.147$
			& $0.278 \pm 0.133$
			& $0.308 \pm 0.148$
			& $0.271 \pm 0.105$
			& $0.302 \pm 0.117$ \\
			&
			& MAC $\uparrow$
			& $0.881 \pm 0.096$
			& $0.855 \pm 0.115$
			& $0.906 \pm 0.092$
			& $0.885 \pm 0.112$
			& $0.916 \pm 0.063$
			& $0.896 \pm 0.077$ \\
			\addlinespace[2pt]
			
			& \multirow{2}{*}{\textbf{TIDAL}}
			& RMSE $\downarrow$
			& $\mathbf{0.237 \pm 0.079}$
			& $\mathbf{0.253 \pm 0.082}$
			& $\mathbf{0.241 \pm 0.079}$
			& $\mathbf{0.259 \pm 0.082}$
			& $\mathbf{0.254 \pm 0.080}$
			& $\mathbf{0.274 \pm 0.084}$ \\
			&
			& MAC $\uparrow$
			& $\mathbf{0.938 \pm 0.042}$
			& $\mathbf{0.930 \pm 0.046}$
			& $\mathbf{0.936 \pm 0.043}$
			& $\mathbf{0.926 \pm 0.047}$
			& $\mathbf{0.929 \pm 0.045}$
			& $\mathbf{0.918 \pm 0.051}$ \\
			
			\bottomrule
		\end{tabular}
	}
\end{table*}

TIDAL consistently achieves the lowest RMSE under both sampling budgets and across all three geometries.
The MAC results show the same overall trend, with TIDAL attaining the highest or closely comparable structural similarity across the evaluated settings.
Together with the H15/V15 results in the main text, these results show that the advantage of the proposed reconstruction framework is maintained across different levels of H/V sampling sparsity.

\section{Complete Analysis of Spatial-Rank Constraints}
\label{app:rank_relaxation}

This section provides additional details for the spatial-rank analysis in Sec.~\ref{sec:rank_analysis}.
We first characterize the singular-value structure of the learned dense spatial components, then examine the representation capacity under prescribed spatial ranks, and finally compare factorized spatial decoders with the dense-\(S\) decoder used in TIDAL.

\subsection{Learned Spatial-Component Spectra}

We analyze the effective rank structure of the learned dense spatial components $S_{\theta,q}\in\mathbb{C}^{200\times200}$.
Across 80 test fields, $Q_N=40$ components, and three sampling geometries, this yields 9,600 matrices.
For each matrix, we compute the cumulative SVD energy
\begin{equation}
	E(r)
	=
	\frac{\sum_{i=1}^{r}\sigma_i^2}
	{\sum_i \sigma_i^2}.
\end{equation}
At each rank $r$, we report the pointwise median and P10--P90 range over all matrices, with all components and sampling geometries weighted equally.

\subsection{Oracle Representation Capacity}
\label{app:oracle_rank}

We evaluate the approximation capacity of the low-rank spatial parameterization
\[
S_q = A_q B_q^\top
\]
for different prescribed spatial ranks $L$.
We fix the number of components to $Q_N=40$ and evaluate
\[
L\in\{2,4,8,16,32,64,100,128,160,200\}.
\]
The experiment uses 20 fields uniformly selected from the test set.
For each field and rank, we run three independent optimizations for 800 steps using Adam with a learning rate of $0.02$, and keep the solution with the lowest reconstruction error.
Statistics are then computed across the 20 fields.
As a reference, we additionally report the corresponding dense-$S$ representation with $Q_N=40$.

\begin{table}[t]
	\centering
	\caption{
		Oracle representation error across different prescribed spatial ranks $L$.
		Results are full-field relative RMSE over 20 test fields.
	}
	\label{tab:oracle_rank_full}
	\small
	\begin{tabular}{lccccc}
		\toprule
		Representation & $L$ & Mean & Median & P10 & P90 \\
		\midrule
		Factorized $S=AB^\top$ & 2   & 0.6020 & 0.6014 & 0.4201 & 0.7084 \\
		Factorized $S=AB^\top$ & 4   & 0.4616 & 0.4514 & 0.2524 & 0.6014 \\
		Factorized $S=AB^\top$ & 8   & 0.3379 & 0.3326 & 0.1188 & 0.5145 \\
		Factorized $S=AB^\top$ & 16  & 0.2518 & 0.2545 & 0.0647 & 0.4130 \\
		Factorized $S=AB^\top$ & 32  & 0.1855 & 0.1815 & 0.0344 & 0.3335 \\
		Factorized $S=AB^\top$ & 64  & 0.1355 & 0.1254 & 0.0256 & 0.2641 \\
		Factorized $S=AB^\top$ & 100 & 0.1140 & 0.1008 & 0.0246 & 0.2340 \\
		Factorized $S=AB^\top$ & 128 & 0.1086 & 0.0967 & 0.0162 & 0.2239 \\
		Factorized $S=AB^\top$ & 160 & 0.0996 & 0.0965 & 0.0032 & 0.2222 \\
		Factorized $S=AB^\top$ & 200 & 0.0992 & 0.0965 & 0.0032 & 0.2222 \\
		\midrule
		Dense $S$ & -- & \textbf{0.0992} & \textbf{0.0965} & \textbf{0.0032} & \textbf{0.2222} \\
		\bottomrule
	\end{tabular}
\end{table}

Table~\ref{tab:oracle_rank_full} shows a clear reduction in representation error as the prescribed spatial rank increases.
Low and moderate ranks retain a substantial approximation gap: the mean RMSE decreases from $0.3379$ at $L=8$ to $0.1855$ at $L=32$ and $0.1355$ at $L=64$, whereas the dense-$S$ reference reaches $0.0992$.
The factorized representation approaches this reference only at very high ranks, with $L=160$--$200$ required to nearly close the gap.
These results indicate that imposing a moderate prescribed spatial rank can introduce a substantial representation bottleneck for the acoustic fields considered here.

\subsection{Effect of Prescribed Spatial Rank in Neural Reconstruction}
\label{app:factorized_neural}

We next examine whether the same limitation appears in end-to-end reconstruction.
We compare factorized spatial decoders with $L=32$ and $L=64$ against the dense-$S$ decoder used in TIDAL.
All models share the same backbone, training data, and optimization settings, differing only in the parameterization of the spatial components.

\begin{table}[t]
	\centering
	\caption{
		End-to-end reconstruction with factorized and dense spatial decoders under the H15/V15 protocol.
		Results are mean relative RMSE over 80 test fields.
		``Full'' and ``Unobs.'' denote full-field and unobserved-region relative RMSE, respectively.
	}
	\label{tab:factorized_neural_full}
	\small
	\resizebox{\textwidth}{!}{
		\begin{tabular}{lccccccc}
			\toprule
			& & \multicolumn{2}{c}{Equispaced}
			& \multicolumn{2}{c}{Shifted}
			& \multicolumn{2}{c}{Jittered} \\
			\cmidrule(lr){3-4}
			\cmidrule(lr){5-6}
			\cmidrule(lr){7-8}
			Decoder & Params
			& Full $\downarrow$ & Unobs. $\downarrow$
			& Full $\downarrow$ & Unobs. $\downarrow$
			& Full $\downarrow$ & Unobs. $\downarrow$ \\
			\midrule
			
			Factorized, $L=32$
			& 21.30M
			& $0.6523$ & $0.6668$
			& $0.6585$ & $0.6752$
			& $0.6682$ & $0.6855$ \\
			
			Factorized, $L=64$
			& 23.27M
			& $0.6452$ & $0.6600$
			& $0.6535$ & $0.6707$
			& $0.6643$ & $0.6823$ \\
			
			\textbf{Dense $S$}
			& \textbf{19.36M}
			& $\mathbf{0.2561}$ & $\mathbf{0.2692}$
			& $\mathbf{0.2594}$ & $\mathbf{0.2744}$
			& $\mathbf{0.2780}$ & $\mathbf{0.2950}$ \\
			
			\bottomrule
		\end{tabular}
	}
\end{table}

Increasing the prescribed spatial rank from $L=32$ to $L=64$ yields only a small improvement in neural reconstruction, reducing the average unobserved-region RMSE across the three geometries from $0.6758$ to $0.6710$.
In contrast, the dense-$S$ decoder achieves an average unobserved-region RMSE of $0.2795$, with the same qualitative advantage under all three sampling geometries.

Notably, this improvement does not result from a larger model: the dense-$S$ configuration contains $19.36$M trainable parameters, compared with $21.30$M and $23.27$M for the $L=32$ and $L=64$ factorized decoders, respectively.
Together with the oracle representation analysis, these results show that a moderate prescribed spatial rank can create a substantial bottleneck in both representation capacity and end-to-end reconstruction.
This supports the use of data-adaptive spatial components in TIDAL while retaining the physics-derived $S/c$ representation.

\section{Complete Data- and Model-Efficiency Results}
\label{app:data_model_efficiency}

This section provides the complete results underlying the data- and model-efficiency analysis in Sec. \ref{sec:efficiency_scaling}. We compare the $S/c$ decoder with a direct-output decoder under different training-set sizes and model widths. Unless otherwise specified, all evaluations use H15/V15 observations and the same 80 test fields under the equispaced, shifted, and jittered sampling geometries.

\subsection{Data Efficiency}
\label{app:data_efficiency}

We first vary the number of training fields while fixing the model width to 384. Both decoder variants use the same encoder, training protocol, normalization, and H/V sampling augmentation. Table~\ref{tab:data_efficiency_full} reports the complete results for $N\in\{50,100,200,500\}$.

\begin{table}[t]
	\centering
	\caption{
		Reconstruction performance under different training-set sizes.
		Results are mean $\pm$ standard deviation over 80 test fields.
		``Full'' and ``Unobs.'' denote full-field and unobserved-region relative RMSE, respectively.
		``Avg. Unobs.'' is the average unobserved-region RMSE across the three sampling geometries.
		Lower is better.
	}
	\label{tab:data_efficiency_full}
	\scriptsize
	\resizebox{\textwidth}{!}{
		\begin{tabular}{clccccccc}
			\toprule
			& & \multicolumn{2}{c}{Equispaced}
			& \multicolumn{2}{c}{Shifted}
			& \multicolumn{2}{c}{Jittered}
			& \\
			\cmidrule(lr){3-4}
			\cmidrule(lr){5-6}
			\cmidrule(lr){7-8}
			Train fields & Decoder
			& Full $\downarrow$ & Unobs. $\downarrow$
			& Full $\downarrow$ & Unobs. $\downarrow$
			& Full $\downarrow$ & Unobs. $\downarrow$
			& Avg. Unobs. $\downarrow$ \\
			\midrule
			
			50
			& Direct output
			& $0.906 \pm 0.021$ & $0.973 \pm 0.023$
			& $0.900 \pm 0.021$ & $0.972 \pm 0.023$
			& $0.902 \pm 0.020$ & $0.975 \pm 0.022$
			& 0.973 \\
			
			& \textbf{$S/c$}
			& $\mathbf{0.876 \pm 0.028}$ & $\mathbf{0.928 \pm 0.026}$
			& $\mathbf{0.871 \pm 0.025}$ & $\mathbf{0.928 \pm 0.022}$
			& $\mathbf{0.872 \pm 0.026}$ & $\mathbf{0.930 \pm 0.024}$
			& \textbf{0.929} \\
			
			\midrule
			
			100
			& Direct output
			& $0.704 \pm 0.076$ & $0.754 \pm 0.082$
			& $0.698 \pm 0.074$ & $0.754 \pm 0.080$
			& $0.714 \pm 0.076$ & $0.771 \pm 0.082$
			& 0.760 \\
			
			& \textbf{$S/c$}
			& $\mathbf{0.641 \pm 0.077}$ & $\mathbf{0.677 \pm 0.078}$
			& $\mathbf{0.632 \pm 0.078}$ & $\mathbf{0.671 \pm 0.080}$
			& $\mathbf{0.650 \pm 0.075}$ & $\mathbf{0.692 \pm 0.076}$
			& \textbf{0.680} \\
			
			\midrule
			
			200
			& Direct output
			& $0.503 \pm 0.096$ & $0.539 \pm 0.103$
			& $0.500 \pm 0.094$ & $0.539 \pm 0.102$
			& $0.525 \pm 0.094$ & $0.566 \pm 0.102$
			& 0.548 \\
			
			& \textbf{$S/c$}
			& $\mathbf{0.396 \pm 0.104}$ & $\mathbf{0.418 \pm 0.108}$
			& $\mathbf{0.395 \pm 0.103}$ & $\mathbf{0.419 \pm 0.107}$
			& $\mathbf{0.419 \pm 0.101}$ & $\mathbf{0.446 \pm 0.105}$
			& \textbf{0.428} \\
			
			\midrule
			
			500
			& Direct output
			& $0.292 \pm 0.088$ & $0.313 \pm 0.094$
			& $0.294 \pm 0.086$ & $0.317 \pm 0.093$
			& $0.317 \pm 0.089$ & $0.343 \pm 0.097$
			& 0.324 \\
			
			& \textbf{$S/c$}
			& $\mathbf{0.256 \pm 0.083}$ & $\mathbf{0.269 \pm 0.085}$
			& $\mathbf{0.259 \pm 0.082}$ & $\mathbf{0.274 \pm 0.085}$
			& $\mathbf{0.278 \pm 0.085}$ & $\mathbf{0.295 \pm 0.089}$
			& \textbf{0.280} \\
			
			\bottomrule
		\end{tabular}
	}
\end{table}

The structured $S/c$ decoder achieves lower full-field and unobserved-region error for every tested training-set size and sampling geometry. The advantage is already present with only 50 training fields and becomes more pronounced with moderate supervision. For example, at $N=200$, the average unobserved-region RMSE decreases from $0.548$ with direct-output decoder to $0.428$ with structured decoding. The same trend is observed independently under equispaced, shifted, and jittered sampling, indicating that the benefit is not specific to a particular H/V geometry.

\subsection{Model Efficiency}
\label{app:model_efficiency}

We next vary the encoder width while using all 500 training fields. The tested widths are $\{128,192,256,384\}$. At each width, the $S/c$ and direct-output models use the same backbone and training protocol, with only a small difference in total parameter count introduced by the corresponding output heads.

\begin{table}[t]
	\centering
	\caption{
		Reconstruction performance under different model capacities.
		Results are mean $\pm$ standard deviation over 80 test fields.
		``Params'' denotes the total number of trainable parameters.
		Lower is better.
	}
	\label{tab:model_efficiency_full}
	\scriptsize
	\resizebox{\textwidth}{!}{
		\begin{tabular}{clcccccccc}
			\toprule
			& & &
			\multicolumn{2}{c}{Equispaced}
			& \multicolumn{2}{c}{Shifted}
			& \multicolumn{2}{c}{Jittered}
			& \\
			\cmidrule(lr){4-5}
			\cmidrule(lr){6-7}
			\cmidrule(lr){8-9}
			Width & Decoder & Params
			& Full $\downarrow$ & Unobs. $\downarrow$
			& Full $\downarrow$ & Unobs. $\downarrow$
			& Full $\downarrow$ & Unobs. $\downarrow$
			& Avg. Unobs. $\downarrow$ \\
			\midrule
			
			128
			& Direct output & 2.17M
			& $0.809 \pm 0.062$ & $0.867 \pm 0.067$
			& $0.808 \pm 0.064$ & $0.872 \pm 0.069$
			& $0.814 \pm 0.061$ & $0.879 \pm 0.066$
			& 0.873 \\
			
			& \textbf{$S/c$} & 2.19M
			& $\mathbf{0.662 \pm 0.081}$ & $\mathbf{0.706 \pm 0.085}$
			& $\mathbf{0.662 \pm 0.079}$ & $\mathbf{0.710 \pm 0.084}$
			& $\mathbf{0.679 \pm 0.075}$ & $\mathbf{0.729 \pm 0.080}$
			& \textbf{0.715} \\
			
			\midrule
			
			192
			& Direct output & 4.85M
			& $0.634 \pm 0.103$ & $0.680 \pm 0.111$
			& $0.633 \pm 0.101$ & $0.683 \pm 0.110$
			& $0.652 \pm 0.099$ & $0.704 \pm 0.107$
			& 0.689 \\
			
			& \textbf{$S/c$} & 4.89M
			& $\mathbf{0.481 \pm 0.106}$ & $\mathbf{0.511 \pm 0.111}$
			& $\mathbf{0.480 \pm 0.103}$ & $\mathbf{0.512 \pm 0.110}$
			& $\mathbf{0.510 \pm 0.105}$ & $\mathbf{0.545 \pm 0.111}$
			& \textbf{0.523} \\
			
			\midrule
			
			256
			& Direct output & 8.57M
			& $0.453 \pm 0.104$ & $0.485 \pm 0.112$
			& $0.453 \pm 0.101$ & $0.489 \pm 0.110$
			& $0.477 \pm 0.103$ & $0.515 \pm 0.111$
			& 0.497 \\
			
			& \textbf{$S/c$} & 8.65M
			& $\mathbf{0.360 \pm 0.102}$ & $\mathbf{0.380 \pm 0.106}$
			& $\mathbf{0.363 \pm 0.100}$ & $\mathbf{0.386 \pm 0.104}$
			& $\mathbf{0.384 \pm 0.100}$ & $\mathbf{0.409 \pm 0.105}$
			& \textbf{0.391} \\
			
			\midrule
			
			384
			& Direct output & 19.20M
			& $0.292 \pm 0.088$ & $0.313 \pm 0.094$
			& $0.294 \pm 0.086$ & $0.317 \pm 0.093$
			& $0.317 \pm 0.089$ & $0.343 \pm 0.097$
			& 0.324 \\
			
			& \textbf{$S/c$} & 19.36M
			& $\mathbf{0.256 \pm 0.083}$ & $\mathbf{0.269 \pm 0.085}$
			& $\mathbf{0.259 \pm 0.082}$ & $\mathbf{0.274 \pm 0.085}$
			& $\mathbf{0.278 \pm 0.085}$ & $\mathbf{0.295 \pm 0.089}$
			& \textbf{0.280} \\
			
			\bottomrule
		\end{tabular}
	}
\end{table}

The $S/c$ decoder also provides consistent gains across model capacities. The improvement is particularly clear for lightweight and intermediate models: at width 192, where both variants contain approximately $4.9$M parameters, the average unobserved-region RMSE decreases from $0.689$ to $0.523$. Similar improvements are observed at widths 128 and 256, while the advantage remains at width 384. Since the paired models have comparable parameter counts at each width, these results indicate that the improvement is associated with the structured output parameterization rather than simply increased model size.

Together, the data- and model-efficiency results show that the $S/c$ decoder remains beneficial across different supervision and capacity regimes, supporting its role as an effective inductive bias for sparse acoustic-field reconstruction.

\end{document}